\documentclass{article} % For LaTeX2e
\usepackage{iclr2026_conference,times}

\usepackage{amsmath,amsfonts,bm}

\def\eqref#1{equation~\ref{#1}}
\def\1{\bm{1}}

\def\mE{{\bm{E}}}

\def\mH{{\bm{H}}}

\def\mL{{\bm{L}}}
\def\mM{{\bm{M}}}

\def\mP{{\bm{P}}}

\def\mR{{\bm{R}}}
\def\mS{{\bm{S}}}

\def\mV{{\bm{V}}}
\def\mW{{\bm{W}}}

\def\mY{{\bm{Y}}}

\DeclareMathAlphabet{\mathsfit}{\encodingdefault}{\sfdefault}{m}{sl}
\SetMathAlphabet{\mathsfit}{bold}{\encodingdefault}{\sfdefault}{bx}{n}

\usepackage{graphicx}
\usepackage{wrapfig}
\usepackage{hyperref}
\usepackage{url}
\usepackage{tcolorbox}
\usepackage{titlesec}
\usepackage{listings}
\usepackage{enumitem}
\usepackage{courier}
\title{Modeling the language cortex with form-independent and enriched representations of sentence meaning reveals remarkable semantic abstractness}

\author{
\textbf{Shreya Saha}$^{1}$\thanks{Correspondence: ssaha@ucsd.edu},  
\textbf{Shurui Li}$^{2}$, 
\textbf{Greta Tuckute}$^{3}$,
\textbf{Yuanning Li}$^{2}$,
\textbf{Ru-Yuan Zhang}$^{4}$, 
\textbf{Leila Wehbe}$^{5}$,\\
\textbf{Evelina Fedorenko}$^{6}$
\textbf{Meenakshi Khosla}$^{1}$\\[4pt]
$^{1}$University of California San Diego, 
$^{2}$ShanghaiTech University, 
$^{3}$Kempner Institute, Harvard University,\\
$^{4}$Shanghai Jiao Tong University, 
$^{5}$Carnegie Mellon University, 
$^{6}$Massachusetts Institute of Technology
}

\iclrfinalcopy % Uncomment for camera-ready version, but NOT for submission.
\begin{document}

\maketitle

\begin{abstract}
% How does the human language cortex encode meaning? We address this question by comparing neural activity during sentence comprehension with representational predictions from vision and language models. When we generate images corresponding to sentences and extract vision model embeddings, we find that aggregating across multiple generated images yields increasingly accurate predictions of language cortex responses—sometimes rivaling large language models. Similarly, averaging embeddings from multiple paraphrases of a sentence improves prediction accuracy compared to any single phrasing. Further, when paraphrases are enriched with commonsense context (e.g., augmenting "I had a pancake" to include implicit details like "maple syrup"), prediction accuracy increases further, surpassing the original sentence. Together, these results provide converging evidence that the meaning component of language cortex responses can be well captured by diverse representational surrogates that abstract away from surface linguistic form.

% When we generate images corresponding to sentence meanings, we find that vision model embeddings for a single (even high-quality) image achieve substantial, non-trivial performance, although their predictive accuracy remains lower than that of the language-model embeddings. Further, we find that aggregating across multiple generated images yields increasingly accurate predictions of language cortex responses, sometimes rivaling large language models.
The human language system represents both linguistic forms and meanings, but the abstractness of the meaning representations remains debated. Here, we searched for abstract representations of meaning in the language cortex by modeling neural responses to sentences  using representations from vision and language models. 
When we generate images corresponding to sentences and extract vision model embeddings, we find that aggregating across multiple generated images yields increasingly accurate predictions of language cortex responses—sometimes rivaling large language models. 
Similarly, averaging embeddings across multiple paraphrases of a sentence improves prediction accuracy compared to any single paraphrase. Enriching paraphrases with contextual details that may be implicit (e.g., augmenting "I had a pancake" to include details like "maple syrup") further increases prediction accuracy, even surpassing predictions based on the embedding of the original sentence, suggesting that the language system maintains richer and broader semantic representations than language models. Together, these results demonstrate the existence of highly abstract, form-independent meaning representations within the language cortex.

\end{abstract}

\section{Introduction}

\begin{figure}[h!]
\centering
\includegraphics[width=\linewidth]{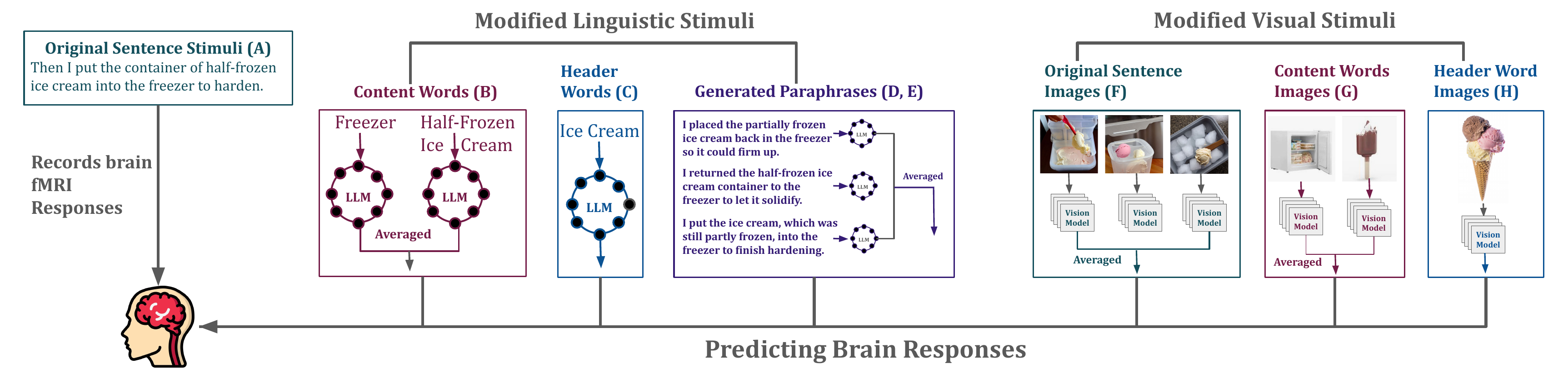}
\caption{We probe whether neural activity in the language cortex can be explained by different representations of the original linguistic input. Starting from each original sentence, we derive and analyze eight alternative representations - (A) original sentence, (B) content words capturing the main semantic elements, (C) a header phrase summarizing a group of related sentences, (D) paraphrases of the original sentence, (E) paraphrase of the original sentence enriched with commonsense context, (F) images generated from the original sentence, (G) images generated from the content words, and (H) images generated from the header phrase. Language-based variants (A–E) are embedded with large language models, while visual variants (F–H) are embedded using vision models.}
\label{fig:schematic}
\end{figure}

% In recent years, large language models have shown remarkable promise in modeling neural activity in the human language cortex. Evidence suggests that the language cortex is sensitive not only to the semantic content of linguistic stimuli but also to the structural properties of language, such as word order and syntax. Although semantic content accounts for the largest share of brain–model similarity, syntactic structure nonetheless plays a critical role. Work comparing sentences with their paraphrases (inputs that preserve meaning but alter form) consistently shows that the original linguistic form provides the strongest prediction of cortical responses. There has been a lot of research focused on disentangling the contributions of different components of linguistic input. In this paper, we analyze how the semantic component of linguistic inputs contributes to brain–model similarity. We ask: (i) to what extent does the original form or modality matter in transmitting semantic information? (ii) Can brain–model similarity be further improved by explicitly enriching the semantic information beyond what is present in the original input?

In recent years, Large Language Models (LLMs) have shown remarkable potential in modeling neural activity in the human language cortex. A central approach for studying these brain–model correspondences has been encoding models, which predict neural responses from features of linguistic stimuli. Early studies relied on corpus derived, automated and hand-constructed feature spaces including word embeddings, phonemes, syntactic structure, or narrative properties distinctions to construct encoding models of the language cortex \cite{mitchell2008predicting, wehbe2014simultaneously,  huth2016natural, de2017hierarchical}. These works demonstrated that word-level representations could account for meaningful variance in brain responses, but often ignored the broader sentence context in which words are embedded. Subsequent efforts showed that incorporating context, capturing relationships between words across time leads to improved modeling of language cortex activity \cite{wehbe2014aligning,jain2018incorporating}. Building on these findings, recent studies have turned to large artificial neural network–based language models, demonstrating that modern LLMs optimized for next-word prediction (e.g., GPT-style models) or trained with contextual objectives (e.g., masked language models such as BERT) provide SOTA predictions of neural responses and reveal striking convergence between artificial and biological language systems \cite{toneva2019interpreting, schwartz2019inducing,schrimpf2020artificial, goldstein2022shared, toneva2022combining, hosseini2024artificial,tuckute2024language}.

Next, researchers have sought to understand which components of language drive the similarity between model representations and brain responses.
A recent study demonstrated that semantic meaning, rather than surface-level lexical or syntactic features, is the dominant factor underlying brain–model alignment \cite{kauf2024lexical}. Research all along has also consistently upheld the importance of preserving the sentence form for activating the language cortex: original sentences reliably yield higher responses than word-level manipulations (e.g., word lists or paraphrases), highlighting the role of meaningful compositional structure. However, to our knowledge, there has been little systematic investigation of whether alternative representations (alternating either in form or even modality) that preserve semantic content can similarly predict language cortex activity.

\textbf{In this paper, we examine whether the information encoded in the human language cortex can be modeled from diverse representational sources that differ substantially from the original linguistic stimulus in both form and modality.} We also explore the role of commonsense knowledge, information that is implicit to humans but not explicitly present in the sentence in shaping cortical representations. Our key contributions are as follows:
% In this paper, our central goal is to examine whether the information encoded in the human language cortex can be modeled from diverse representational sources, even when these differ substantially from the original linguistic stimulus, both in form and in modality. We ask whether brain–model similarity is specific to the original linguistic sentences presented to participants during neural recordings, or whether it extends to alternative representations that preserve or enrich the underlying semantic content. We also explore the role of commonsense knowledge—information that is implicit to humans but not explicitly present in the sentence—in shaping cortical representations.
%While these models may not achieve a perfect match, they demonstrate that the language cortex can be meaningfully modeled not only from direct linguistic inputs but also from diverse representations that convey equivalent or enriched semantic content. Furthermore, we show that augmenting inputs with commonsense knowledge—information that is readily available to humans but not explicitly stated in the input sentence—substantially improves prediction accuracy. 

\begin{enumerate}[leftmargin=20pt,labelsep=0.5em]
    \item We demonstrate that embeddings from vision foundation models, applied to visual depictions of sentences, possess non-trivial predictive power for modeling language cortex activity during sentence comprehension. Importantly, this predictivity increases when incorporating multiple diverse images per sentence, demonstrating that representational averaging provides a closer approximation of the format of linguistic meaning in the brain.
    
    %Importantly, this predictivity increases as we incorporate a more diverse set of images for each sentence. This suggests that multiple images can capture different facets of a sentence’s meaning that a single image fails to represent. Our findings highlight that while visual representations alone cannot fully account for language cortex activity, they provide a compelling and complementary approximation of linguistic meaning in the brain.
    
    \item Next, we demonstrate that foundational large language model embeddings of paraphrased sentences also predict language cortex activity, with accuracy improving as more paraphrases are used. This mirrors the visual domain findings: greater representational diversity enhances prediction accuracy through averaging, indicating that the language cortex can be modeled even when stimulus form differs substantially from the original input.

    \item Finally, we show that enriching sentences with commonsense context-information evident to humans but not explicitly present in the original text produces a substantial boost in predictivity. This finding underscores the critical role of implicit background knowledge in shaping cortical representations and suggests that effective brain-model alignment requires integrating structured, contextually relevant knowledge beyond surface-level representations.
\end{enumerate}

\section{Training and Datasets}
\label{subsec:training_dataset}

\begin{figure}[h!]
\centering
\includegraphics[width=0.9\linewidth]{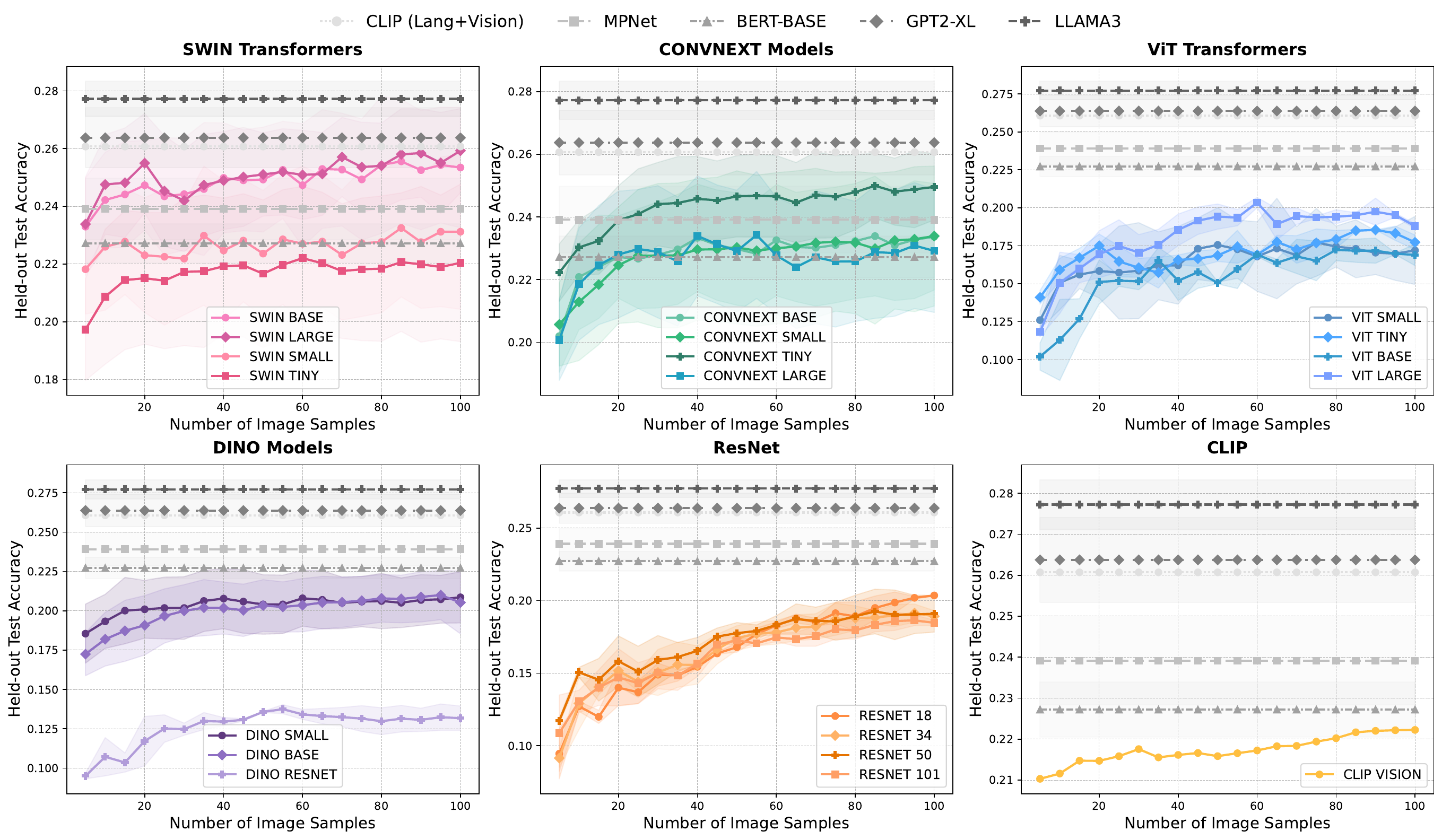}
\caption{Comparison of performance in predicting language cortex activity between LLM embeddings of the original linguistic stimuli from the Pereira (2018) dataset (CLIP, MPNet, BERT-BASE, GPT2-XL and LLAMA3 presented in grey horizontal lines) and vision model embeddings of their corresponding visual counterparts (remaining models). Performance of the visual models increases with the number of images, sometimes surpassing some of the language models.}
\label{fig:Pereira_img_sent}
\end{figure}

We analyzed voxel responses from the `core' language network, comprising the Inferior Frontal Gyrus, Inferior Frontal Gyrus – Orbital part, Middle Frontal Gyrus, Anterior Temporal cortex, and Posterior Temporal cortex across three datasets where participants read and processed sentences during fMRI scanning: Pereira (2018) \cite{pereira2018toward}, Tuckute (2024) \cite{tuckute2024driving} and Caption Scene Dataset CSD (2025) \cite{li2025large}. Additional details on datasets are provided in Appendix \ref{subsec:datasets}.

We first investigate whether meaning in the language cortex is captured exclusively by language model embeddings of the original sentence, or whether it can also be represented by alternative sentences that convey the same information. We also test whether representations from other modalities, such as vision, can capture this meaning (Figure \ref{fig:schematic}). Finally, we assess whether predictive accuracy based on the original sentence can be further improved by enriching it with commonsense contextual information. 

To evaluate predictive accuracy, we fit a ridge regression model $\mY \approx \mE \mW$ by minimizing 
$\|\mY - \mE \mW\|_F^2 + \lambda \|\mW\|_F^2$, 
where $\mY$ denotes the voxel responses, $\mE$ the embeddings from either vision or language models, and $\mW$ the regression weights (examples in Appendix Figures \ref{fig:Pereira_Dataset_Samples}, \ref{fig:content_word_examples}, \ref{fig:header_examples}, \ref{fig:pereira_paraphrases_examples} and \ref{fig:tuckute_paraphrases_examples}). The parameter $\lambda > 0$ is the regularization hyperparameter, chosen via cross-validation to prevent overfitting. More information on training can be found in Appendix Section \ref{subsec:training_details}. 
% We evaluate the models on held-out data by computing XXX (correlation?) over the language network and averaging (?). 
We use the following featurizations:

\begin{enumerate}[leftmargin=20pt,labelsep=0.5em]
    \item[A.] \textbf{Original sentences:} We use the original sentence $\{ S_i \}_{i=1}^N$ presented to the subjects, obtain the penultimate layer embeddings from an LLM as $\mE = \{\mP\mL(S_i) \}_{i=1}^N$ where $\mP\mL(\cdot)$ denotes the penultimate layer embedding.

     \item[B.] \textbf{Content words:} We extract the most concrete and semantically salient terms $\{c_j \}_{j=1}^{C^S}$ from each sentence: open-class parts of speech such as nouns and main verbs carrying the core meaning of the sentence. They are identified using a SciPy-based syntactic parser. For example, from ``The boy is eating pancakes'', we extract ``boy'' and ``pancakes''. Each content word is embedded separately using the penultimate layer of the LLM, then averaged to create a sentence-level representation:  $\mE = \frac{1}{C_S} \Sigma_{c_j} \mP\mL(c_j)$.

      \item[C.] \textbf{Header words:}
      For the Pereira (2018) dataset \cite{pereira2018toward}, sentences are grouped into paragraphs sharing a common topic. We use the paragraph header $\mH$ as a high-level semantic summary. The header embedding $\mE = \mP\mL(\mH)$ is used to predict the averaged brain responses $\mY = \frac{1}{K} \Sigma_{j=1}^K \mY_{S_j}$ across all sentences in the paragraph, where $\mY_{S_j}$ is the brain response for sentence $S_j$ and $K$ is the number of sentences in the paragraph. Note that in the Pereira (2018) dataset’s third experiment, several paragraphs described the same topic, we therefore merged all such paragraphs and used the shared topic as the paragraph header.

       \item[D.] \textbf{Standard paraphrases:} We generate $\mR=70$ paraphrases (Appendix Table \ref{tab:para_prompt}) for each sentence using Gemini \cite{comanici2025gemini} (alternative phrasings that preserve the original semantic meaning while varying in surface form). To analyze how the number of paraphrases affects prediction accuracy, we systematically sample subsets in increments of 5 (i.e., $r\in\{5,10,15,.,70\}$). For each subset size $r$, we average the embeddings $\mE = \frac{1}{r} \Sigma_{i=1}^r \mP\mL(P_i^S)$, where $P_i^S$ is the $i^{th}$ paraphrase generated for sentence $\mS$. This incremental sampling allows us to track how sentence-level representational diversity impacts brain prediction accuracy.

       \item[E.] \textbf{Enriched paraphrases:} We generate $\mR=70$ paraphrases (Appendix Table \ref{tab:para_context_prompt}) that embed broader contextual and inferential content: details that a human might naturally associate with a sentence, even if not explicitly stated. We process these enriched paraphrases in the same way as the standard paraphrases described in D. 
                    
    \item[F.] \textbf{Sentence-Generated Images:} We use the stable diffusion model \cite{rombach2022high} to generate $M=100$ images for each sentence $\mS$, using the sentence itself as the prompt. While generating textual descriptions from images is relatively well-studied, expressing the full meaning of a sentence in a single image is far more challenging. Sentences often convey abstract concepts or temporally extended events that cannot be captured in one snapshot. To mitigate this, we create a diverse set of images per sentence, each offering a complementary visual perspective that together approximate the sentence’s semantics. From these images we extract embeddings from the penultimate layer of a state-of-the-art vision model $\{ \mV\mM(I_i^S) \}_{i=1}^M$. We then average a subset of $m\in\{5,10,15,..,100\}$ out of $M$ embeddings to form a single representation $E = \frac{1}{m}\Sigma_{i=1}^m \mV\mM(I_i^S)$, where $I_i^S$ is the $i^{th}$ image generated for sentence $\mS$.
    
    \item[G.] \textbf{Content-word images:} For each content word $\{c_j \}_{j=1}^{C^S}$ in sentence $\mS$, we use the stable diffusion model to generate $M$ images. From these images we extract embeddings from the penultimate layer of a state-of-the-art vision model $\{ \mV\mM(I_k^{c_j}) \}_{k=1}^M$. We first average a selected subset of $m$ embeddings to obtain a single representation for each content word, and then average across all content word representations to produce a sentence-level embedding $\mE = \frac{1}{C^S} \Sigma_{c_j} \frac{1}{m} \Sigma_{k=1}^m \mV\mM(I_k^{c_j})$, where $I_k^{c_j}$ is the $k^{th}$ image generated for content word $c_j$.
    
    \item[H] \textbf{Header images:} We generated $M$ images of the same header word, get vision model embeddings of them, and average a subset $m$ of these images to get an input representation $\mE = \frac{1}{m} \Sigma_{i=1}^m \mV\mM(I_i^H)$ to predict $\mY = \frac{1}{K} \Sigma_{j=1}^K \mY_{S_j}$ as described above, where $I_i^H$ is the $i^{th}$ image generated for header word $\mH$.

\end{enumerate}

We used a wide range of vision models spanning diverse architectures and training objectives \cite{he2016deep, liu2022convnet, liu2021swin, dosovitskiy2020image, caron2021emerging, radford2021learning}, and language models spanning encoder–decoder and decoder-only architectures, with causal and non-causal variants 
\cite{song2020mpnet, devlin2019bert,radford2019language, team2025gemma, dubey2024llama} (details are provided in Appendix Sections \ref{subsec:lang_details} and \ref{subsec:vision_details}).

\section{Results}
\subsection{Visual Models Capture Meaning in the Language Cortex}
%\subsection{Vision model embeddings derived from visual descriptions of linguistic stimuli show strong potential for modeling responses in the language cortex}

\subsubsection{Sentence-level comparison}
%\subsubsection{Comparison of LLM embeddings of entire sentences with vision model embeddings of their visual counterparts}
\label{subsec:img_sent}

% \begin{wrapfigure}{l}{0.5\linewidth} % r = right side; change to l for left
%   \centering
%   \includegraphics[width=\linewidth]{images/csd_single_img.pdf}
%   \caption{Performance comparison between SOTA LLM embeddings of the original linguistic stimuli from the CSD 2025 dataset and single SOTA vision model embedding of their original COCO visual counterparts.}
%   \label{fig:csd_single_img}
% \end{wrapfigure}

% \begin{wrapfigure}{r}{0.5\linewidth} % r = right side; change to l for left
%   \centering
%   \includegraphics[width=\linewidth]{images/pereira_single_image.pdf}
%   \caption{Performance of vision models in predicting language cortex activity
%   using a single image, with experiments repeated across images sorted in
%   order of decreasing quality.}
%   \label{fig:pereira_single_image}
% \end{wrapfigure}

\begin{figure}[ht]
  \centering
  %--- Left figure ---
  \begin{minipage}[t]{0.48\linewidth}
    \centering
    \includegraphics[width=\linewidth,height=6cm,keepaspectratio]{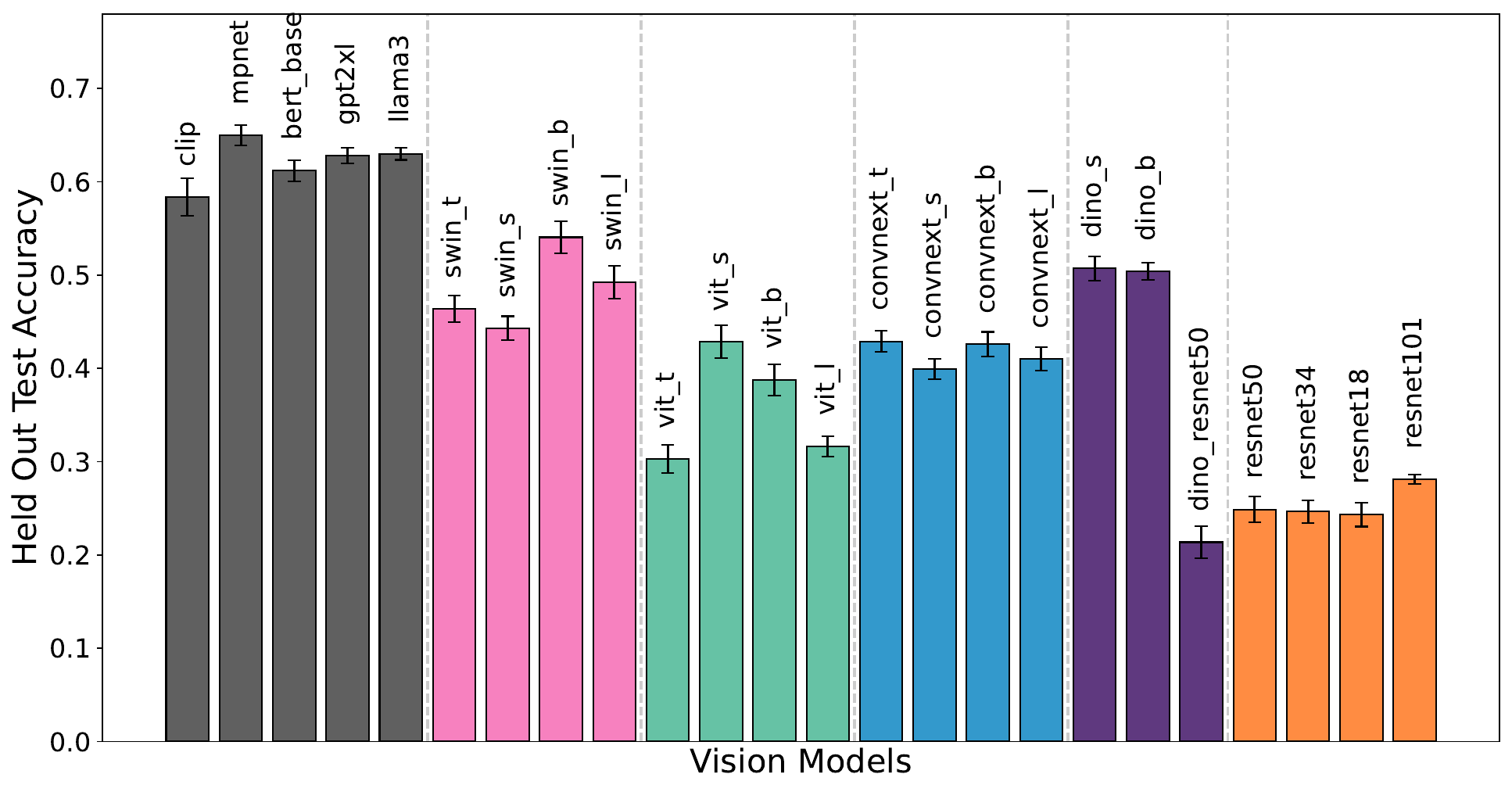}
    \caption{Comparison of performance in predicting language cortex activity between LLM embeddings of the original
    linguistic stimuli from CSD (2025) dataset (first 5 bars) and single image vision model
    embedding of their original COCO visual counterparts (remaining bars).}
    \label{fig:csd_single_img}
  \end{minipage}
  \hfill
  %--- Right figure ---
  \begin{minipage}[t]{0.48\linewidth}
    \centering
    \includegraphics[width=\linewidth,height=6cm,keepaspectratio]{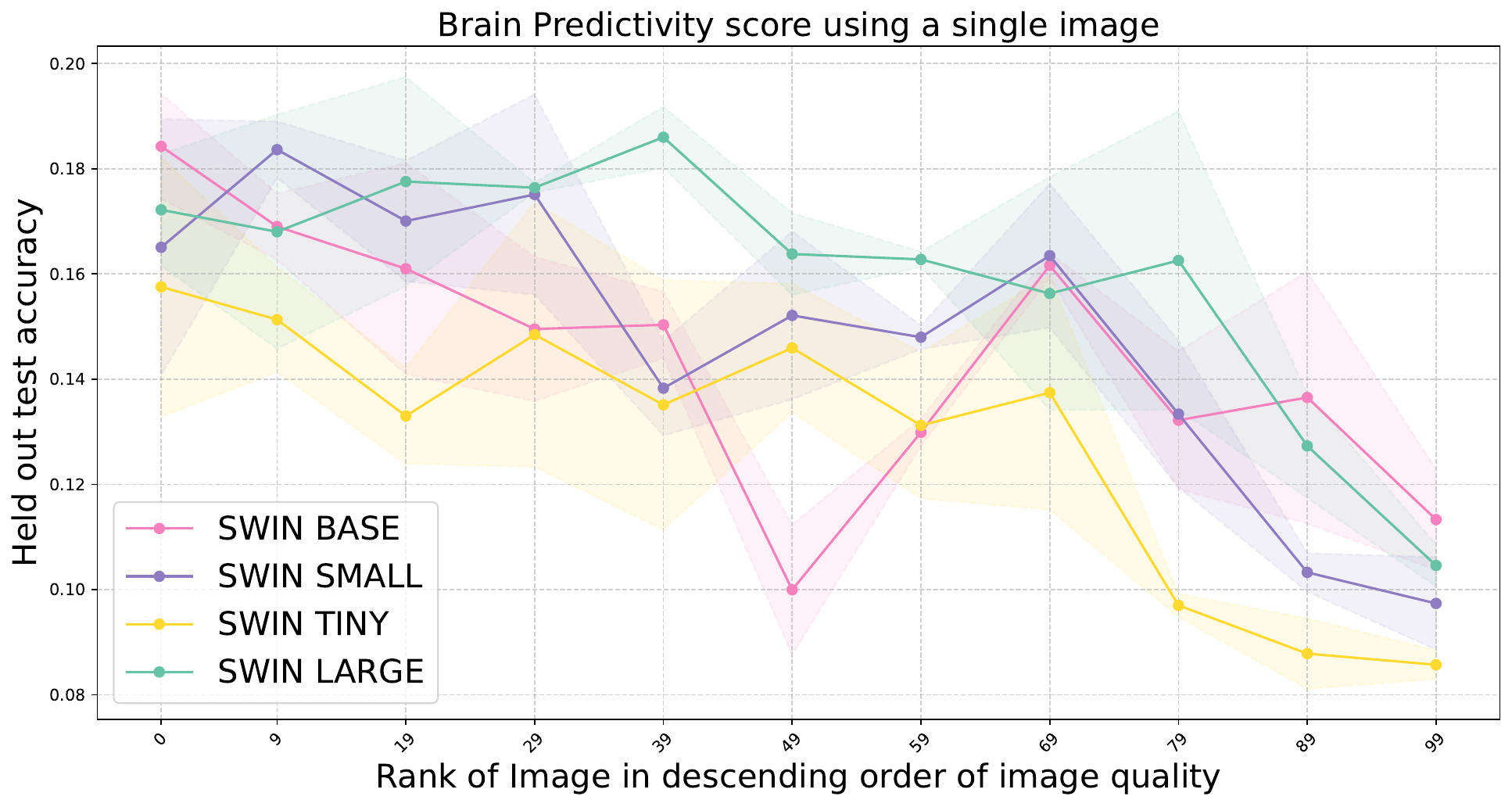}
    \caption{Performance of SWIN vision models in predicting language cortex activity
    using a single image, with experiments repeated across images sorted in
    order of decreasing quality (defined as the cosine similarity between the sentence and the image’s CLIP embeddings).}
    \label{fig:pereira_single_image}
  \end{minipage}
\end{figure}

First, we compare how vision and language models predict brain responses in the language cortex. Using LLM embeddings of each full sentence and vision-model embeddings of their corresponding generated images (A vs F), we train encoding models to predict cortical activity. With only a single image per sentence, language-based models outperform vision-based ones, though vision embeddings also demonstrate meaningful predictive power for language cortex activity, with some models (such as SWIN transformers) achieving competitive performance. This pattern emerges both in the Pereira (2018) dataset using images generated from sentence prompts (Figure \ref{fig:Pereira_img_sent}) and in the CSD (2025) dataset using original COCO images that correspond to the caption stimuli (Figure \ref{fig:csd_single_img}). As we increase the number of generated images per sentence and average their embeddings, the performance of vision models improves substantially—sometimes even surpassing certain language models (see Pereira (2018) results in Figure \ref{fig:Pereira_img_sent}, Tuckute (2024) results in Appendix Figures \ref{fig:greta_img_sent_good_clip}, and CSD (2025) in Appendix Figure \ref{fig:csd_img_sent}). 

This cross-modal success demonstrates that language cortex responses can be predicted using representations from visual content that preserves the semantic meaning. While individual images may miss certain semantic nuances, aggregating multiple visual perspectives systematically improves neural prediction accuracy. The robustness of this representational averaging effect, where performance gains from multiple images occur consistently across diverse vision architectures and training objectives highlights that this is a fundamental property of how semantic content can be distilled from varied visual exemplars, even though baseline performance levels differ across model families.  

% This pattern indicates that a single image rarely captures the full semantic richness of a sentence, whereas a diverse set of images can approximate its meaning more effectively. These findings support the view that sentence-level semantics can be reconstructed, at least in part, through visual representations when those representations encompass a broad range of visual interpretations. This effect is particularly pronounced in the Pereira dataset and holds consistently across a wide range of vision models, regardless of architectural class and training objectives, highlighting the generality and robustness of the phenomenon.

A critical nuance in our analysis is in the quality of the generated images, which is computed as the cosine similarity between CLIP-based embeddings of each image and its corresponding sentence. We use this similarity as a proxy for semantic alignment. Despite being generated from the same sentence, images vary widely in how accurately they capture the intended meaning, as diffusion models are not perfect and often produce off-topic or visually noisy samples (Appendix Figure \ref{fig:Image_Quality}).

Currently, all generated images are treated equally during the averaging process. However, only a subset of these images truly reflect the core content of the sentence, while others may diverge significantly. Our analysis shows that this variance directly impacts model performance: the predictive performance of models using single image embeddings declines systematically with decreasing image quality (Figure \ref{fig:pereira_single_image}), demonstrating that low quality images fail to encode the critical semantic information present in the original sentence, leading to poorer alignment with brain responses.

\begin{figure}[h!]
\centering
\includegraphics[width=0.9\linewidth]{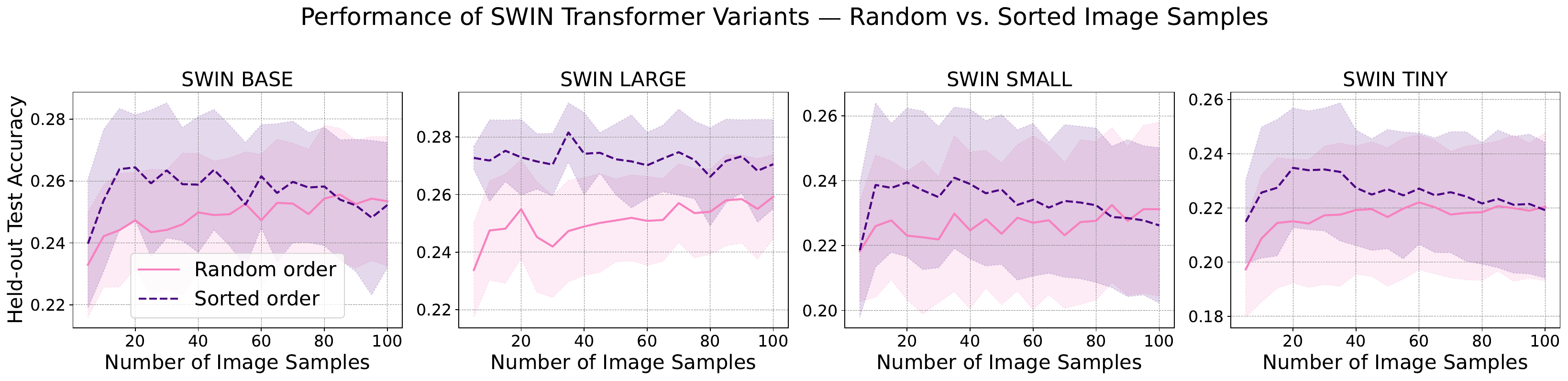}
\caption{Pereira (2018) Dataset - Comparison of vision model performance in predicting language cortex activity using multiple images, with images ordered randomly versus in order of decreasing quality. Averaging more images helps in the case of random ordering, consistent with averaging the noise from non-ideal images. Adding more images eventually hurts in the case of ordered images, where eventually, less useful images are being incorporated.}
\label{fig:pereira_sorted_images}
\end{figure}

We tested whether sorting images by semantic quality yielded more interpretable performance trajectories. When we incrementally include images in descending order of CLIP-based semantic similarity, we observe a sharp rise in accuracy as high-quality images are added, followed by a plateau, and eventually a decline as lower-quality images introduce noise (Figure \ref{fig:pereira_sorted_images}). This contrasts with random ordering, where prediction accuracy is consistently lower and continually increases with exemplar averaging, confirming that image quality systematically affects neural prediction accuracy.

\subsubsection{Content-word comparison}
%Comparison of LLM embeddings of content words of a sentences with vision model embeddings of their visual counterparts}

\begin{figure}[ht]
  \centering
  % --- Left figure ---
  \begin{minipage}[t]{0.48\linewidth}
    \centering
    \includegraphics[width=0.9\linewidth,height=5cm,keepaspectratio]{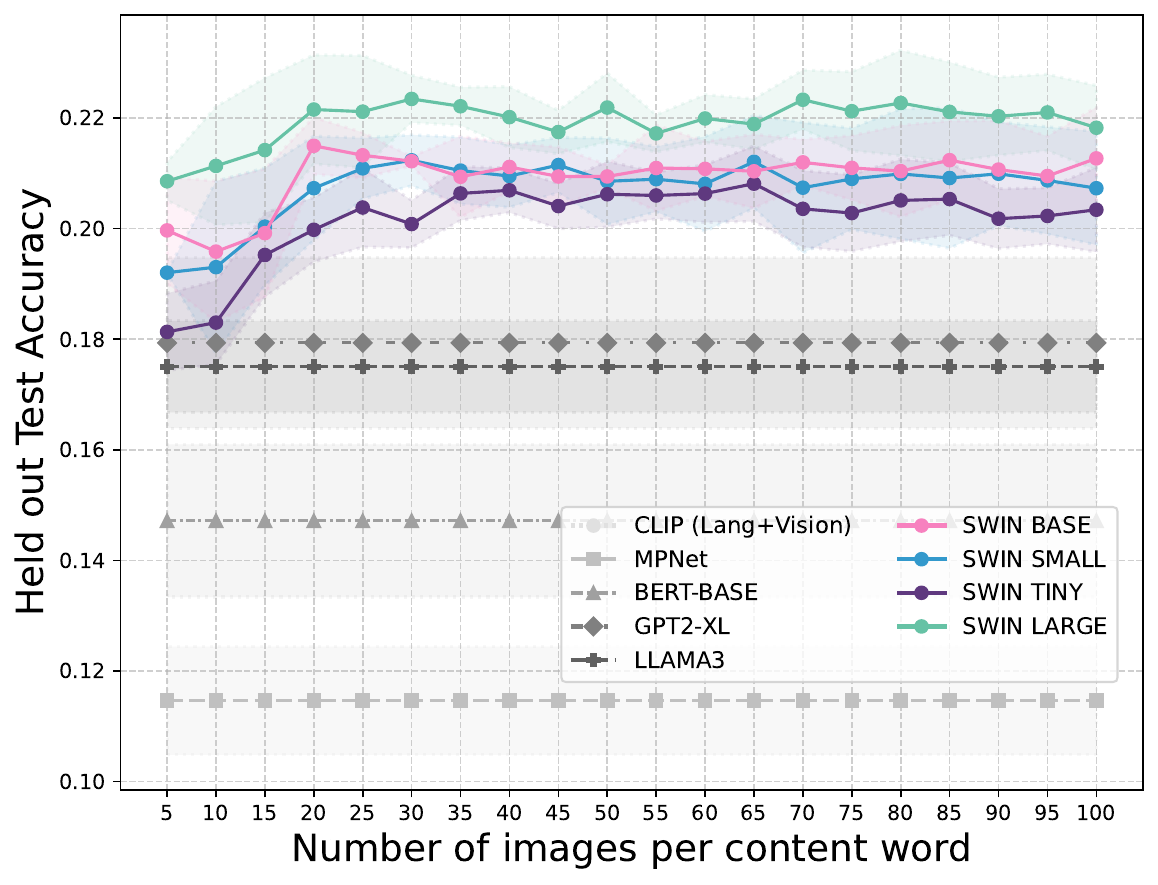}
    \caption{Comparison of performance in predicting language cortex activity between LLM embeddings of the content-word
    linguistic stimuli and SWIN model embeddings of their
    visual counterparts. Performance of SWIN vision models (which are trained without language supervision) increases with the number of images, surpassing many of the language models.}
    \label{fig:content_words}
  \end{minipage}
  \hfill
  % --- Right figure ---
  \begin{minipage}[t]{0.48\linewidth}
    \centering
    \includegraphics[width=0.9\linewidth,height=5cm,keepaspectratio]{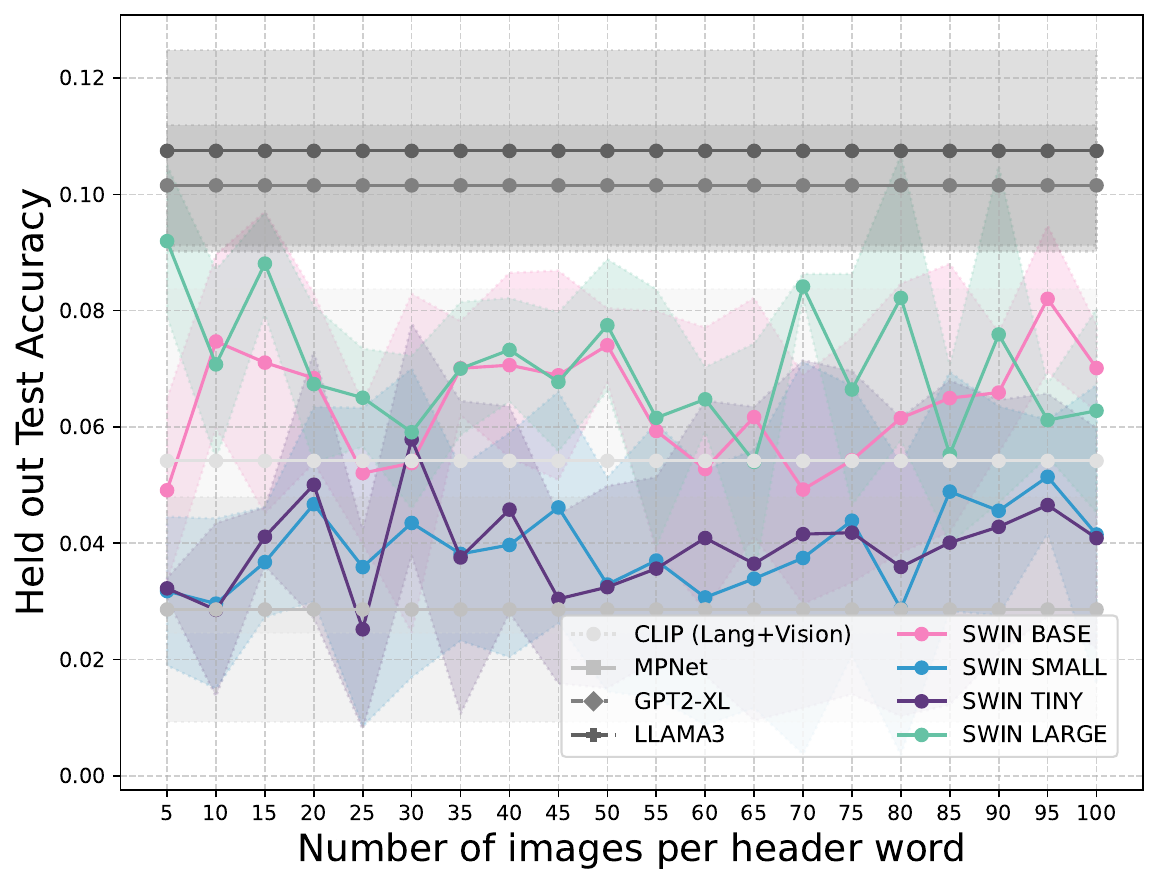}
    \caption{Comparison of performance at predicting language cortex activity between  LLM embeddings of the header word
    groups from the Pereira (2018) dataset and SWIN model embeddings of
    their visual counterparts. For both language and vision models, the performance is worse than when looking at content-words or the original sentence. }
    \label{fig:header_words}
  \end{minipage}
\end{figure}

Next, we tested whether the core semantic elements of sentences could predict language cortex responses when stripped of their full compositional structure. We evaluated the capacity of vision and language models to predict language-cortex responses by focusing on content words within each sentence—primarily nouns and main verbs that carry the essential semantic content (examples in Appendix Figure \ref{fig:content_word_examples}). This approach reduces sentences to their key conceptual building blocks while removing grammatical structure, function words, and compositional relationships. For each content word, we extracted embeddings from language models, and for the corresponding generated images produced by Stable Diffusion, we obtained visual embeddings using vision networks (featurization B vs G) (Figure \ref{fig:content_words}). Even with a small number of content-word-based images, vision models achieved performance levels comparable to, and in most cases surpassing, those of certain language models such as BERT-Base and MPNet. We also observed a consistent upward trend in accuracy with increasing numbers of images, suggesting that aggregating multiple visual representations of individual concepts captures semantic nuances that contribute to language cortex responses.

% \begin{wrapfigure}{r}{0.5\linewidth}
% \centering
% \includegraphics[width=\linewidth]{images/content_words.pdf}
% \caption{Performance comparison between SOTA LLM embeddings of the content words original linguistic stimuli and SOTA vision model embeddings of their corresponding visual
% counterparts.}
% \label{fig:content_words}
% \end{wrapfigure}

% Next, for the Tuckute 2024 dataset (Figure \ref{fig:content_words}), we restricted our analysis to 760 sentences for which content words could be reliably extracted and whose corresponding generated images had an average CLIP similarity score greater than 0.25. While the same trend of performance improvement with more images was present, it was generally weaker than in the Pereira dataset and absent for the Swin-Tiny model. However, the trend did hold for the larger Swin variants. This discrepancy can be attributed to the inherently more abstract nature of the stimuli in the Tuckute 2024 dataset, where feature concept words are less visually concrete compared to those in Pereira. Consequently, vision models struggled more on this dataset and performed significantly worse than language models overall. Nonetheless, they still demonstrated promising potential in modeling language cortex responses.

% \begin{wrapfigure}{r}{0.5\linewidth} % r = right side; change to l for left
%   \centering
%   \includegraphics[width=\linewidth]{images/header_words.pdf}
%   \caption{Performance comparison between SOTA LLM embeddings of the header word comprising of a group of sentences from the Pereira 2018 dataset and SOTA vision model embeddings of their corresponding visual
% counterparts.}
%   \label{fig:header_words}
% \end{wrapfigure}

These experiments reveal that when we reduce sentences to their core conceptual content, vision and language models achieve comparable predictive power for language cortex responses. The fact that vision model embeddings of images of individual words like `boy' and `pancakes' can predict brain responses nearly as well as linguistic representations is particularly striking given that these vision models were trained purely on natural image statistics without language supervision. This suggests that the meaning component of language cortex responses taps into conceptual representations that can be accessed through the statistical structure of visual experience alone, indicating that semantic processing in the language cortex may be grounded in modality-independent principles that both linguistic and visual systems can converge upon. However, the remaining performance differences between content words and full sentences (compare Figures~\ref{fig:content_words} and \ref{fig:Pereira_img_sent}) highlight that compositional structure, grammatical relationships, and contextual integration also contribute meaningfully to language cortex responses, though perhaps to a lesser degree than the core conceptual content.

% These experiments reveal that the performance gap between vision and language models narrows significantly when both modalities have access to comparable semantic information, as demonstrated in the Pereira 2018 dataset. However, when language representations are more semantically rich: capturing abstract, contextual, or relational nuances that are difficult to visualize, vision models tend to lag behind, highlighting the limitations of static visual representations in fully capturing linguistic meaning.

\subsubsection{Header-word comparison}
We now compare LLM embeddings of header words with vision model embeddings of corresponding images to predict averaged brain responses to sentence groups sharing the same topic (C vs H). Vision and language models achieve remarkably comparable performance levels, though both perform substantially lower than when using original sentences or content words (Figure \ref{fig:header_words}). The averaging effect with increasing numbers of images is consistent in smaller Swin models but modest, possibly because individual header words are already concrete and semantically focused enough that a few images  capture their core meaning.
This experiment represents the most abstract test of cross-modal prediction: using single thematic words (e.g., `toaster', `beekeeping') to predict language cortex responses to entire passages on those topics. The near-equivalence between vision and language models at this level suggests that when semantic content is highly abstracted, modality differences become minimal. However, the overall lower accuracy indicates that much of the predictive power for language cortex responses comes from the richer semantic and compositional information present in full sentences and individual concepts, rather than abstract thematic categories alone.
% Lastly we compare LLM header word embeddings and vision model embeddings of their visual counterparts for predicting averaged brain responses to sentences belonging to the same passage (C vs H). We find that increasing the number of images per header word leads to improved performance of the vision models (Figure \ref{fig:header_words}). This effect is especially consistent in smaller Swin models, whereas the trend is somewhat noisier and less reliable in larger Swin variants, echoing similar trends observed in our earlier experiments. It’s worth noting that these results are inherently noisier, as predictions are based on voxel responses averaged across multiple sentence-level stimuli within each paragraph.

% This experiment further equalizes the semantic content available to language and vision models. We began by comparing entire sentence embeddings to their corresponding images, then moved to more aligned comparisons using individual content words and their images. By now abstracting to a shared thematic representation via header words used in both textual and visual forms, we take a further step in bridging the gap between modalities. These findings further highlight the promise of vision models in modeling the language cortex, even when provided with extremely abstract and minimal input, they are still able to capture meaningful structure in the brain responses.

\subsection{Paraphrases}

\subsubsection{Comparing Original Sentences with Paraphrases}

\begin{figure}[h!]
\centering
\includegraphics[width=0.9\linewidth]{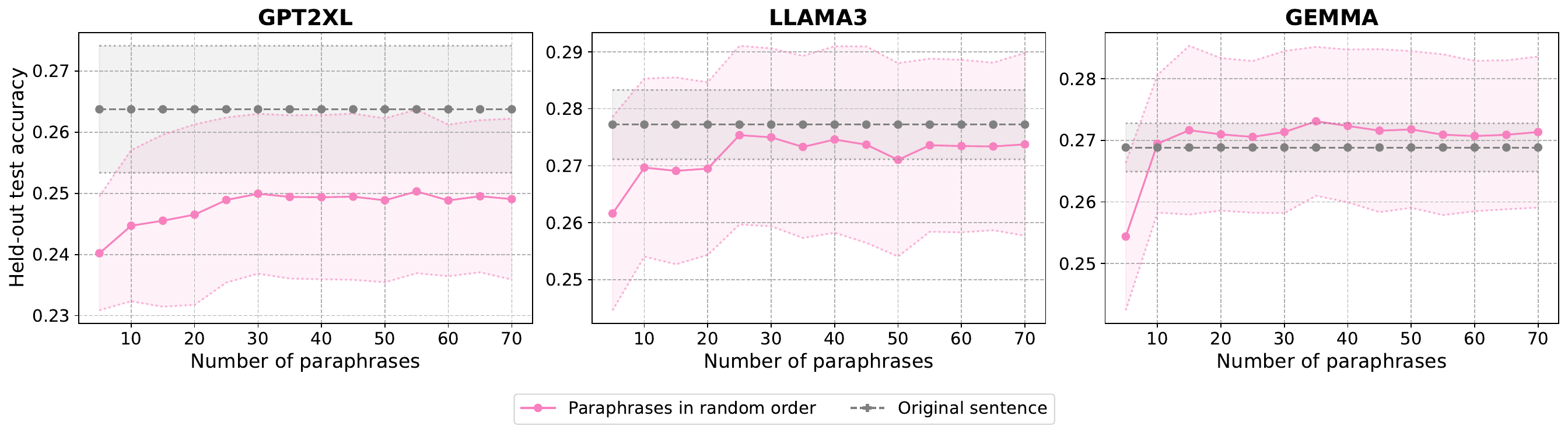}
\caption{Pereira (2018) dataset - Comparison of the performance in predicting language cortex activity by LLM embeddings of the original linguistic stimuli presented to subjects with averaged embeddings of generated paraphrases. Averaging embeddings across more paraphrases steadily improves prediction accuracy, in some cases matching or surpassing the original sentence.}
\label{fig:paraphrase_no_pereira_org}
\end{figure}

So far, we examined the potential of cross-modal representations for modeling the language cortex. Here, we turn to alternative representations within the same modality, comparing the modeling performance of LLM embeddings of the original sentence with the mean embeddings of its paraphrases (A vs D). In the Pereira (2018) dataset, paraphrase embeddings alone yielded reasonable prediction accuracy.  While averaging over a small number of paraphrases ($\leq 5$) underperformed relative to the original sentence embeddings (Figure \ref{fig:paraphrase_no_pereira_org}), performance improved steadily as more paraphrases were incorporated (pink curve), at times surpassing the original sentence baseline (grey curve). This benefit was weaker in the Tuckute (2024) dataset (Appendix Figure \ref{fig:paraphrase_no_tuckute_org}), likely for two reasons: (i) responses in this dataset were averaged across voxels, which eliminated voxel-specific information that may carry important signal, and (ii) Pereira (2018) paragraphs are longer, richer and more content-diverse, making paraphrases more likely to be diverse, whereas Tuckute (2024) sentences are simpler and more abstract, yielding paraphrases with limited variation (Appendix Tables \ref{fig:pereira_paraphrases_examples}, \ref{fig:tuckute_paraphrases_examples}).

%Across the Pereira dataset, we find a consistent pattern with all LLMs: averaged embeddings of a few (5 or less) paraphrases predicts brain responses less accurately than embeddings of the original sentence (Figure \ref{fig:paraphrase_no_pereira_org}). However, averaging the embeddings of multiple paraphrases steadily improves performance (pink curve), sometimes beating the performance of the original sentence (grey plot). This effect is weaker in the Tuckute  (Appendix Figure \ref{fig:paraphrase_no_tuckute_org}), likely for two reasons: (i) the signal is noisier because we predict an averaged response across voxels rather than individual voxels, and (ii) While sentences in the Pereira dataset are broader and richer in content,  making paraphrases more likely to introduce additional meaningful context, the Tuckute dataset contains simpler, more abstract sentences, resulting in paraphrases that exhibit less variation. (Appendix Tables \ref{fig:pereira_paraphrases_examples} and \ref{fig:tuckute_paraphrases_examples}).

This pattern parallels our vision-based experiments, where a single image was less predictive than the original sentence but prediction accuracy improved as multiple visual samples were averaged. As in the vision case, alternative modalities or representational forms show greater potential for modeling the language cortex when the original stimuli presented to subjects are information-rich and concrete. Similarly, when we sort the paraphrases based on their semantic similarity to the original sentence (measured using the cosine similarity in the CLIP language encoder space) and incrementally add them in descending order of similarity (Appendix Figures \ref{fig:paraphrase_no_pereira_rand_sort} and \ref{fig:paraphrase_no_tuckute_rand_sort}), we observe a similar trend: accuracy improves rapidly at first and then saturates (green plot). This suggests diminishing returns as lower quality or less semantically similar paraphrases are added, highlighting that not all paraphrases contribute equally, and the quality of information matters as much as quantity.

Averaging paraphrase embeddings narrows, but typically does not erase the gap to the original sentence; it exceeds the original mainly with stronger LLMs (Gemma3) and for semantically rich stimuli (e.g., Pereira (2018)). This pattern suggests two components in the language cortex signal: (i) a content-dominant component recoverable from many surface realizations, and (ii) a form-sensitive residual captured best by the original sentence.

%Nonetheless, even when we average multiple paraphrase embeddings, the resulting accuracy is usually quite close to that of the original sentence (with paraphrases surpassing it only for the most powerful LLMs such as Gemma3). This aligns with prior work suggesting that while paraphrases may retain the same meaning, subtle differences in linguistic form still carry brain-relevant signals. That is, even when semantic content is held constant, the way that content is expressed: its structure or modality can meaningfully impact neural responses.

We also tested whether paraphrases add complementary information to the original by concatenating the original embedding with the paraphrase average (orange curve in Appendix  Figs. \ref{fig:paraphrase_no_pereira_rand_conc}, \ref{fig:paraphrase_no_tuckute_rand_conc}). In Pereira (2018), concatenation surpasses the original, indicating that paraphrases contribute additional, brain-relevant content beyond surface form. In Tuckute (2024), concatenation improves over paraphrases alone but does not surpass the original, consistent with paraphrases adding little new information for short/abstract stimuli. These findings suggest that paraphrase-based representations contribute most when they introduce substantive semantic variation, whereas in the absence of such variation the precise linguistic form of the original sentence remains the strongest predictor of language-cortex responses.

%Next, we asked whether incorporating paraphrases could make an original sentence more predictive of brain responses. We concatenated the original embedding with the averaged paraphrase embedding (orange curve in Appendix Figures \ref{fig:paraphrase_no_pereira_rand_conc} and \ref{fig:paraphrase_no_tuckute_rand_conc}). In the Pereira dataset, accuracy exceeded that of the original sentence, reflecting the richness of context in its paraphrases. In contrast, although there was a substantial improvement over using only paraphrases in the Tuckute dataset, performance did not surpass that of the original sentence, as the paraphrases generated for these more abstract sentences added relatively little new information. Thus, these results underscore that the form of the original stimuli plays a critical role in modeling the language cortex. This aligns with prior work suggesting that, although paraphrases retain the same meaning, subtle differences in linguistic form can carry brain-relevant signals.

\subsubsection{Comparing Original Sentences with Commonsense-Enriched Paraphrases}
\label{subsec:para_details}
\begin{figure}[h!]
\centering
\includegraphics[width=0.9\linewidth]{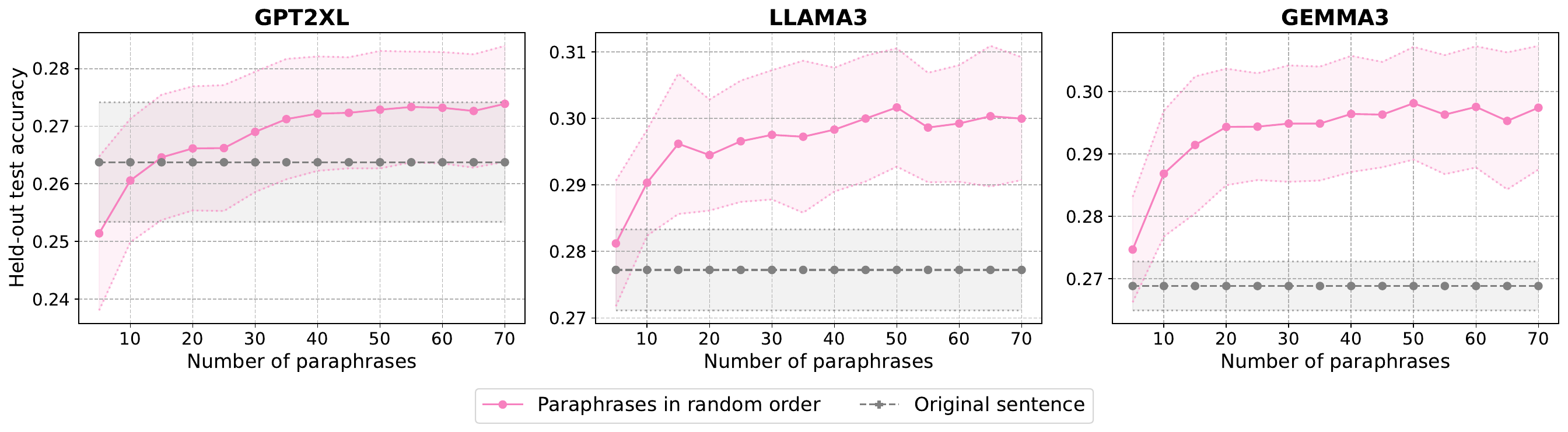}
\caption{Pereira (2018) Dataset - Comparison of LLM embeddings of the original linguistic stimuli presented to subjects with averaged embeddings of generated paraphrases with additional context. Averaging embeddings of paraphrases enriched with contextual information yields higher prediction accuracy than the original sentence embeddings.}
\label{fig:paraphrase_extra_pereira_org}
\end{figure}
Our earlier experiments hinted at a key idea: adding supplemental information to a sentence, beyond its original form can often enhance our ability to model brain  responses. In the above analyses, we preserved the original sentence embedding and concatenated it with embeddings from paraphrases. However, those paraphrases were semantically near-identical to the original sentence, merely differing in surface structure. As a result, the added information was limited, more a stylistic variation than a substantive expansion of meaning.

This led us to ask a deeper question: can paraphrases enriched with new commonsense context, while still preserving the core semantics of the original sentence further improve brain predictivity (featurization A vs E)? In this next set of experiments (illustrated in Figure \ref{fig:paraphrase_extra_pereira_org}, Appendix figure \ref{fig:stat_analysis_paraphrases}), we intentionally moved beyond superficial rewording and generated paraphrases that embedded broader contextual and inferential content: details that a human might naturally associate with the sentence, even if not explicitly stated. These enriched paraphrases preserve the original intent, but vary more significantly in both form and informational density. More broadly, we were interested in whether adding such rich context could compensate for or even outweigh the importance of original form or modality when modeling responses in the language cortex (Appendix Table \ref{fig:pereira_paraphrases_examples}). 

We found that the effectiveness of this approach depended strongly on the nature of the dataset. In the Pereira (2018) dataset, whose sentences are typically rich, concrete, and highly visualizable, it was relatively easy to generate extended paraphrases with coherent, plausible context. In these cases, even a small number of enriched paraphrases substantially outperformed the original sentences in modeling brain activity (pink curve in Figure \ref{fig:paraphrase_extra_pereira_org}), and performance continued to improve without plateauing. This suggests that additional meaningful information, particularly information the brain may implicitly infer can have a strong additive effect on neural predictivity. We next asked whether combining the original sentence with these enriched paraphrases could further improve model performance, reasoning that the specific linguistic form of the original sentence might add information beyond paraphrased content (Appendix Figure \ref{fig:paraphrase_extra_pereira_rand_conc}). When we concatenated the original sentence embedding with the averaged paraphrase embedding, we did not see as major an improvement as we had seen when we used paraphrases without additional context. For older or less powerful models, such as GPT-2 XL, concatenation yielded a modest boost relative to paraphrases alone, likely because these models benefit from the precise structure of the original sentence combined with the additional semantic variation from the paraphrases. However, for more powerful LLMs, such as Llama 3 and Gemma 3, the improvement was negligible, and in some cases the paraphrases alone performed slightly better than the concatenated representation. These stronger models already encode rich, contextually robust representations of the original sentence, so the explicit addition of the original embedding contributes little new information. 

By contrast, for the Tuckute (2024) dataset this strategy was less effective. These sentences are short (6 words long) and some of them abstract, making it difficult to generate paraphrases that are both semantically coherent and genuinely informative. Many paraphrases introduced noise or drifted from the original meaning (Appendix Table \ref{fig:tuckute_paraphrases_examples}), so the enriched paraphrases alone failed to outperform the original sentences (Appendix Figure \ref{fig:paraphrase_extra_tuckute_org}). In this setting, the exact linguistic form of the stimulus carried more predictive power than additional semantic content. As a result, appending the original sentence by concatenation led to performance gains, although it does not always beat the prediction accuracy when the embedding of the original sentence is used. (Appendix Figure \ref{fig:paraphrase_extra_tuckute_rand_conc}).

These results suggest that predictive accuracy in the language cortex is shaped strongly by the informativeness of semantic content. When added information is meaningful and consistent with the brain’s associative priors, it enhances predictivity. Thus, enriching inputs with inferential context can provide substantial gains beyond what is available from the original linguistic form alone.
%When the added information is meaningful and aligns with the brain’s expectations and associative priors, it significantly boosts model predictivity. This emphasizes the importance of semantic completeness in addition to syntactic fidelity: enriching an input with commonsense or inferential context may be more valuable than preserving the original form alone.
\section{Discussion}

In this work, we investigated whether the human language cortex can be modeled by stimuli that differ from the original linguistic input in modality or form while preserving semantic content. Our central aim was to examine the limits of representational flexibility, that is, the extent to which neural responses to linguistic stimuli can be accurately predicted using representations derived from alternative inputs that vary dramatically in surface form (paraphrases), modality (visual vs. linguistic), and information density (enriched vs. original content). 

These analyses provide converging evidence that the human language cortex maintains highly abstract, form-independent representations of semantic meaning. This conclusion emerges from three key observations: vision model embeddings can predict language cortex responses when aggregated across multiple images, paraphrase embeddings show similar predictive power when averaged, and semantically enriched paraphrases can exceed the predictive accuracy of original sentences.

The success of cross-modal prediction using visual representations to model language cortex responses is particularly striking given that the language cortex was not exposed to any visual input during the original experiments. This suggests that meaning in the language cortex is encoded in a format that transcends specific sensory modalities. Further, the effectiveness of representational averaging across examples sharing semantic content: whether multiple images or paraphrases suggests that this process amplifies shared meaning while reducing noise from surface-level variations. This finding demonstrates that the meaning component of language cortex responses can be captured even when input content is embedded in highly variable surface forms or accessed through entirely different sensory modalities. Additionally, the finding that enriched paraphrases can exceed original sentence predictions suggests that the language cortex constructs enriched semantic representations that extend far beyond the information literally present in the text, incorporating the contextual associations and commonsense inferences that humans naturally bring to comprehension. This finding suggests a core difference in the breadth of semantic associations that brains and language models maintain, with the brain capturing richer and more encompassing representations, perhaps due to the broader range of real-world tasks it learns to perform that go beyond next word prediction. 

Several limitations should be acknowledged. First, our visual generation relied on current diffusion models, which may not optimally capture the visual content most relevant to language processing. Future work could explore whether more sophisticated multimodal models or human-generated images yield different patterns. Second, our commonsense enrichment was generated algorithmically and may not reflect the specific knowledge integration processes that occur during natural reading or listening, where representations evolve on a word-by-word basis. Lastly, the fMRI datasets used in this study have inherently slow temporal resolution, providing only coarse snapshots of neural activity averaged over several seconds rather than millisecond-level dynamics. The dataset-dependent effects we observed highlight the need for more diverse neuroimaging corpora that span the full range of linguistic content, from concrete to abstract. 

Despite these limitations, our work demonstrates how powerful generative models can be leveraged to create alternative input types that preserve semantic content while varying in modality (visual vs. linguistic) and surface form (paraphrases), enabling cross-modal and cross-form neural prediction. Through averaging model representations of multiple generated variants, we can isolate the shared meaning components that drive neural responses. This methodological approach opens new avenues for understanding how intelligent systems, both biological and artificial, extract and represent meaning from experience that can be captured across diverse sensory modalities and surface forms.

\bibliography{iclr2026_conference}
\bibliographystyle{iclr2026_conference}

\appendix

\section{Appendix}

\begin{figure}[h!]
\centering
\includegraphics[width=\linewidth]{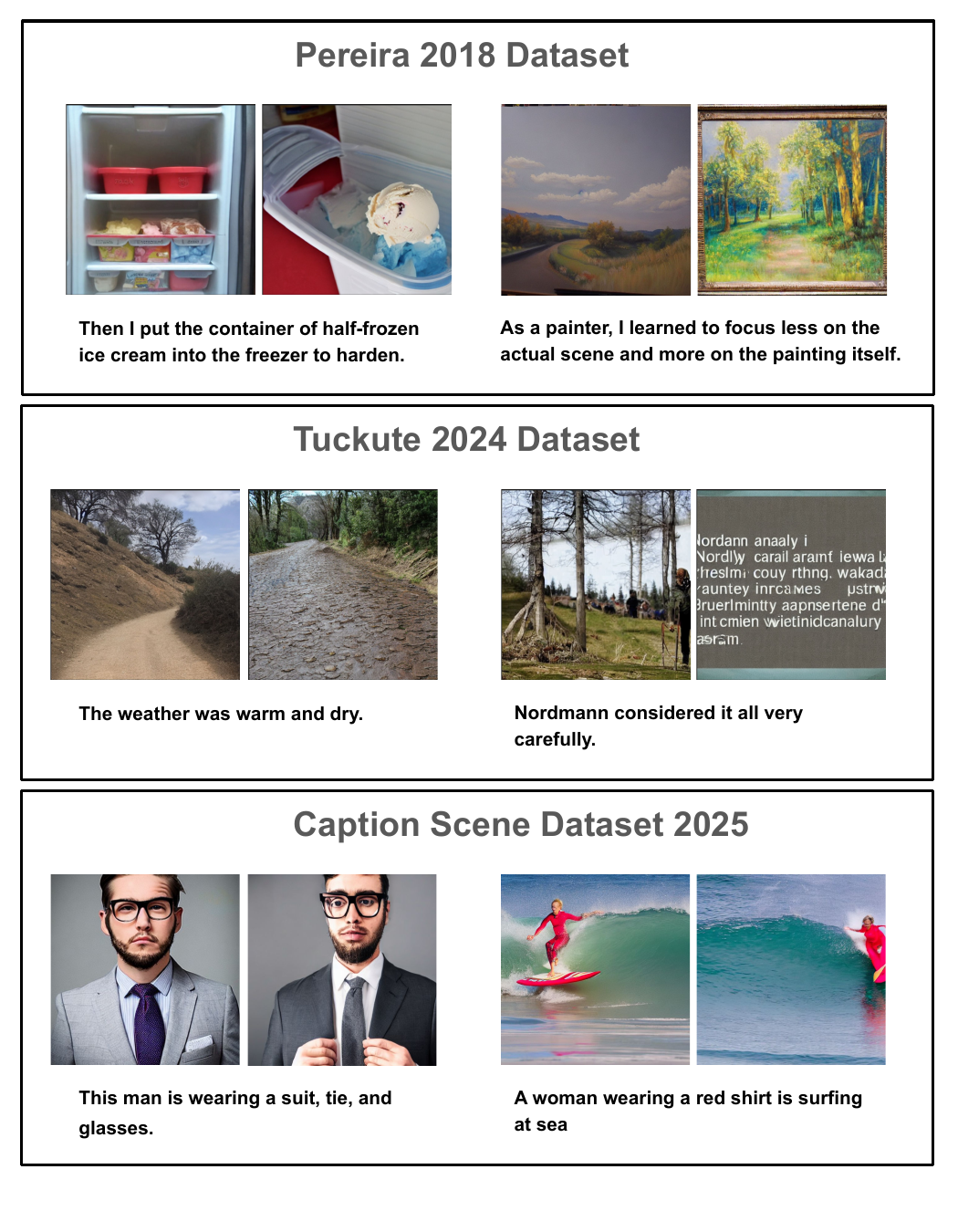}
\caption{Examples of sentences used, and images generated for these sentences using Stable Diffusion}
\label{fig:Pereira_Dataset_Samples}
\end{figure}

\begin{figure}[h!]
\centering
\includegraphics[width=\linewidth]{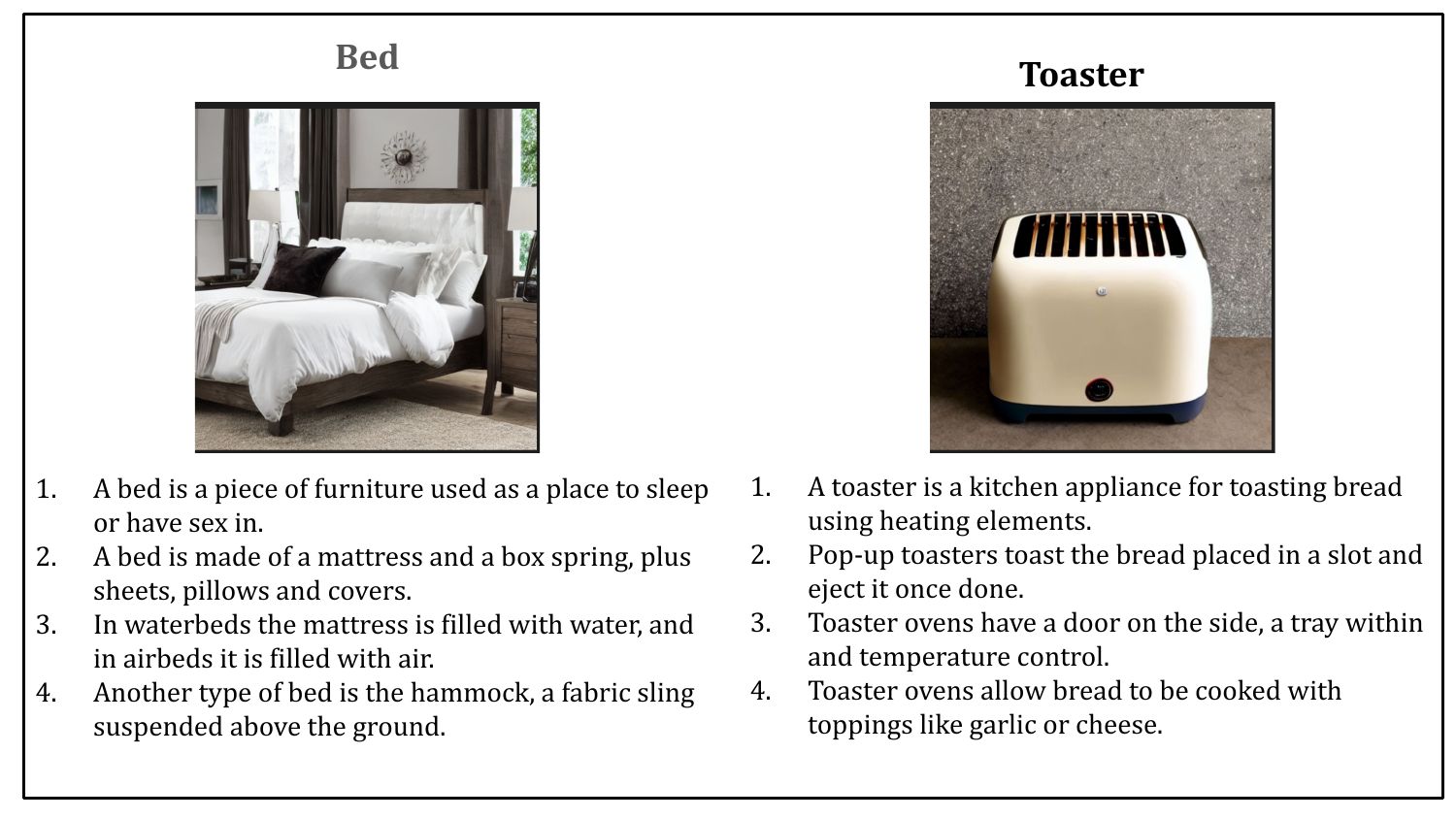}
\caption{Examples of header words for a group of sentences in Pereira (2018) dataset, and images generated for these header words using Stable Diffusion}
\label{fig:header_examples}
\end{figure}

\begin{figure}[h!]
\centering
\includegraphics[width=\linewidth]{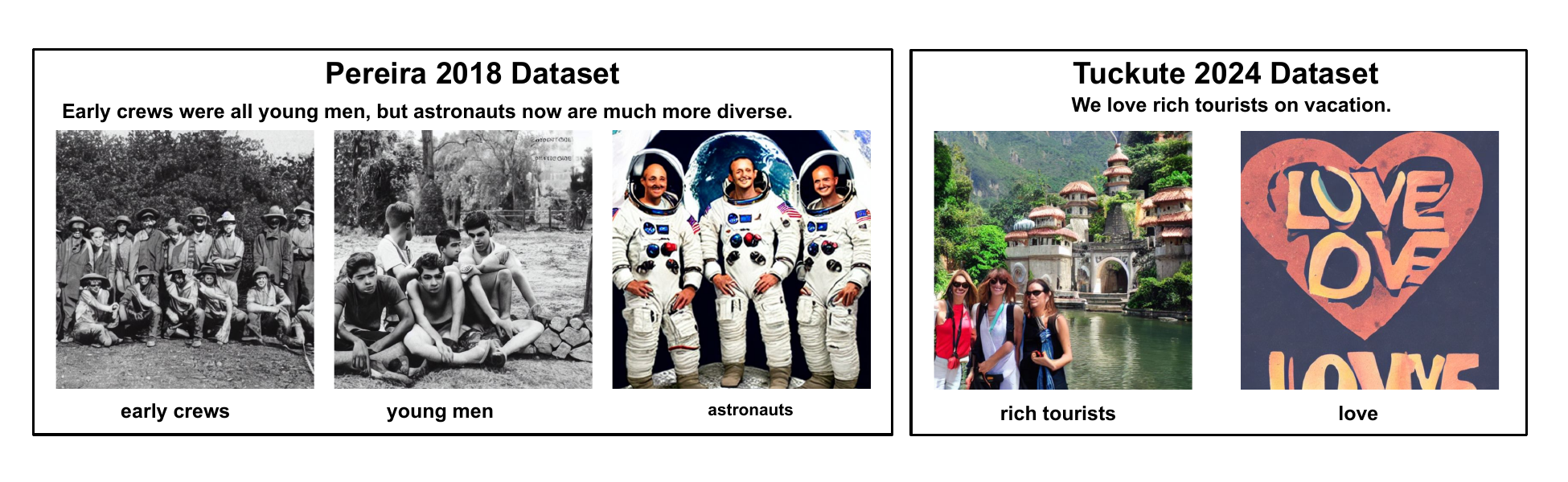}
\caption{Examples of content words/phrases in a sentence, and images generated for these content words/phrases using Stable Diffusion}
\label{fig:content_word_examples}
\end{figure}

\begin{figure}[h!]
\centering
\includegraphics[width=\linewidth]{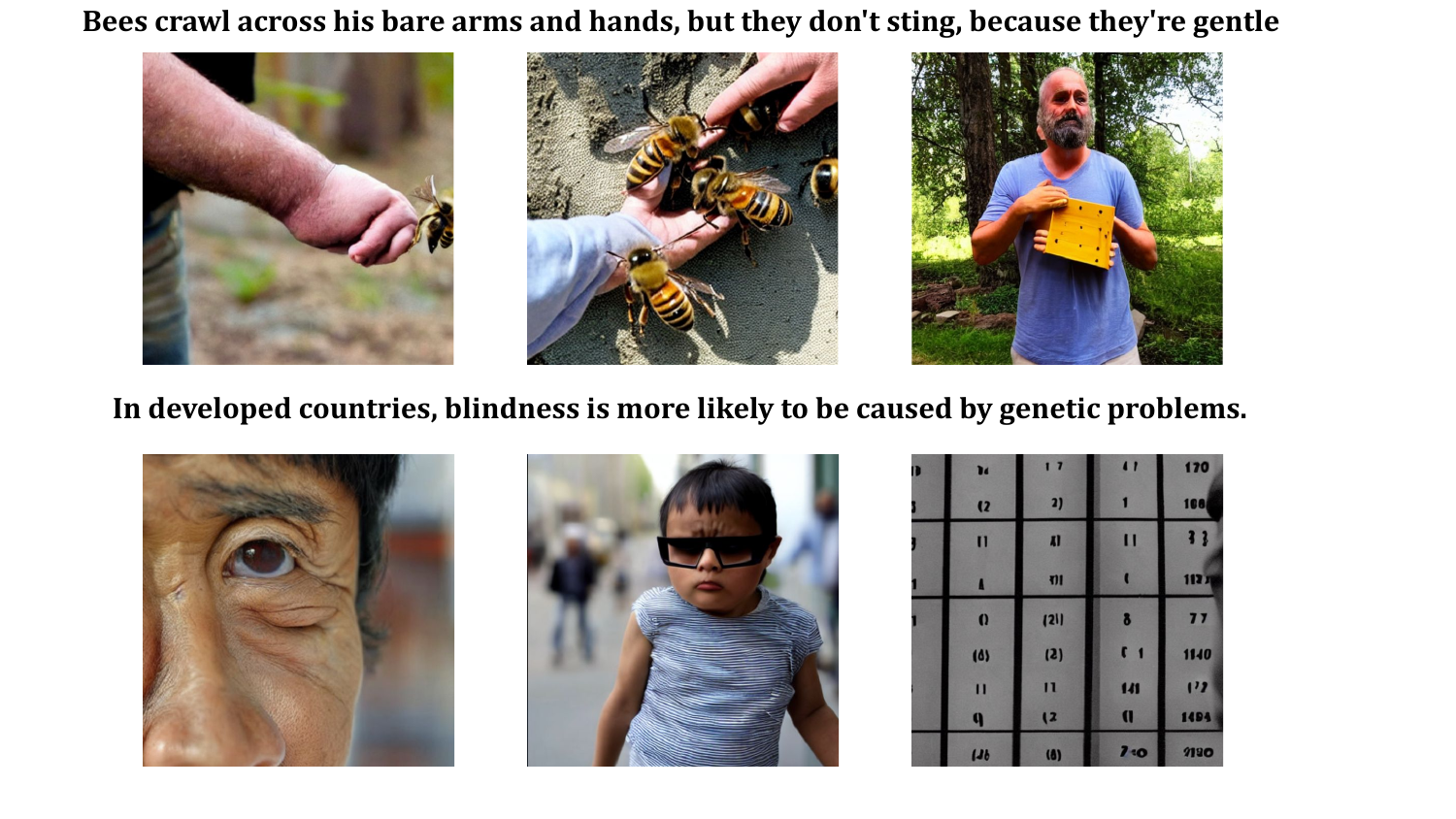}
\caption{Examples of images generated for a given sentence, ranked in descending order of quality. Image quality is measured by the cosine similarity between the CLIP embeddings of the original sentence and the generated images.}
\label{fig:Image_Quality}
\end{figure}

\begin{figure}[h!]
\centering
\includegraphics[width=\linewidth]{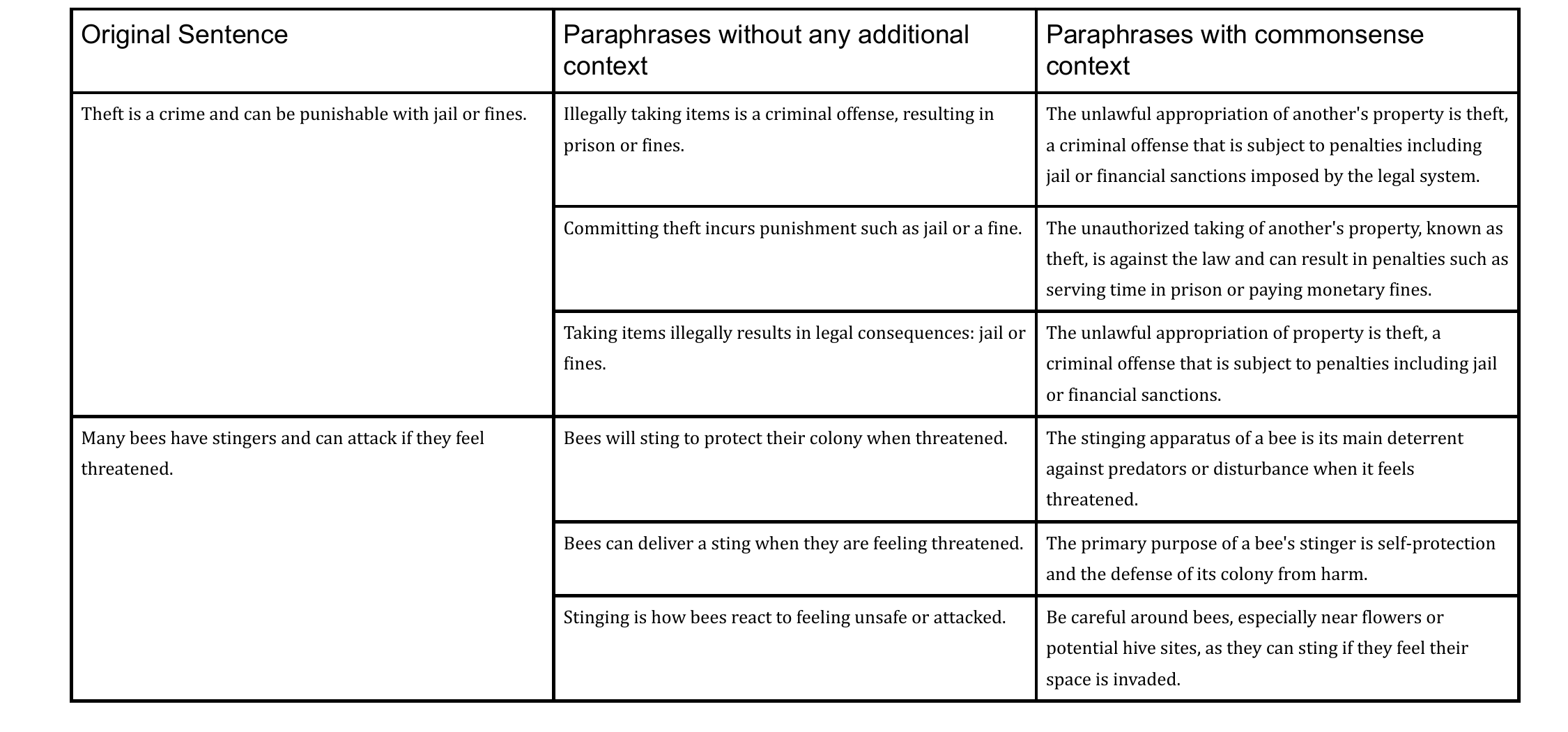}
\caption{Examples of paraphrases generated for sentences in the Pereira (2018) dataset}
\label{fig:pereira_paraphrases_examples}
\end{figure}

\begin{figure}[h!]
\centering
\includegraphics[width=\linewidth]{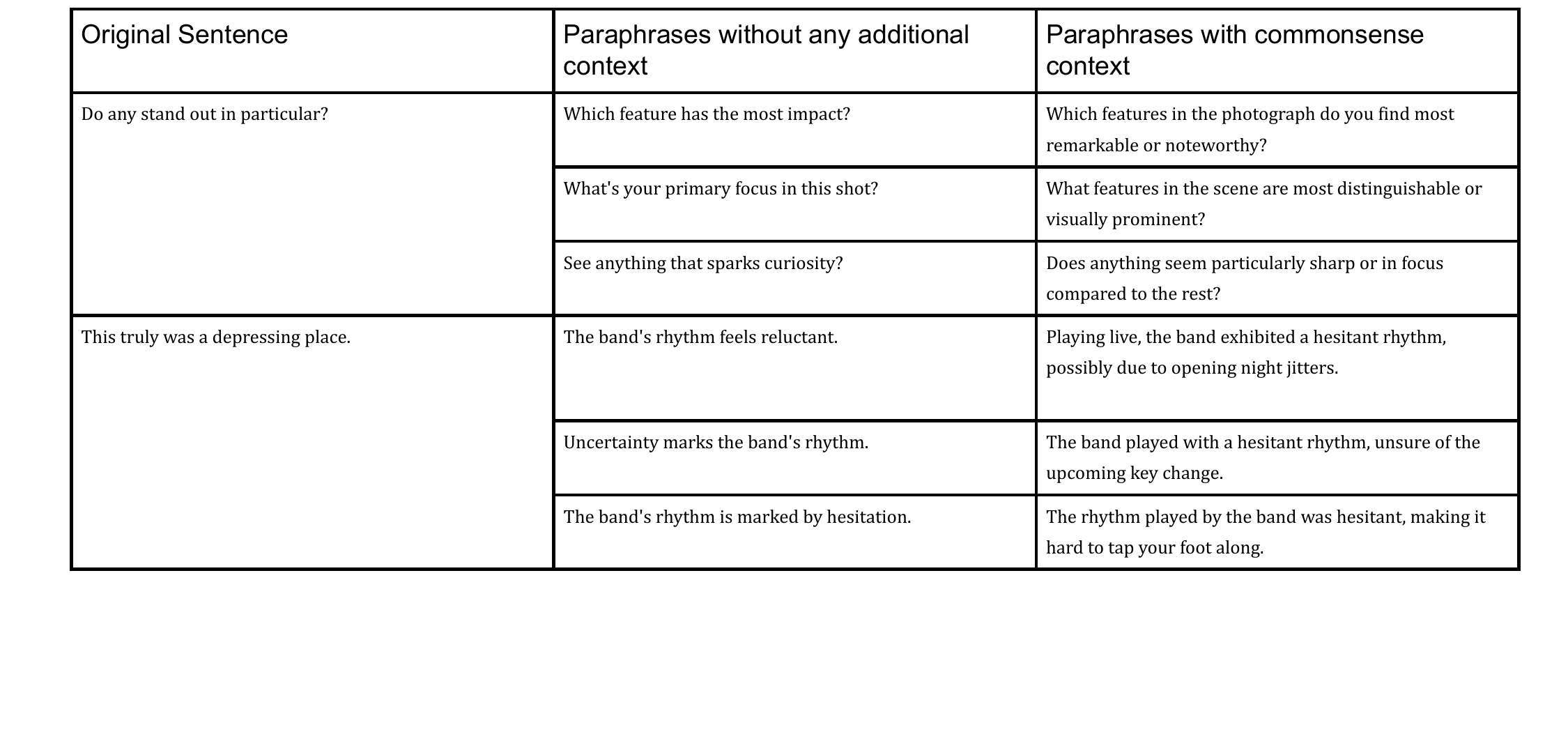}
\caption{Examples of paraphrases generated for sentences in the Tuckute (2024) dataset}
\label{fig:tuckute_paraphrases_examples}
\end{figure}

% \subsection{Comparison of various vision models from vision-lang comparisons}

\begin{figure}[h!]
\centering
\includegraphics[width=\linewidth]{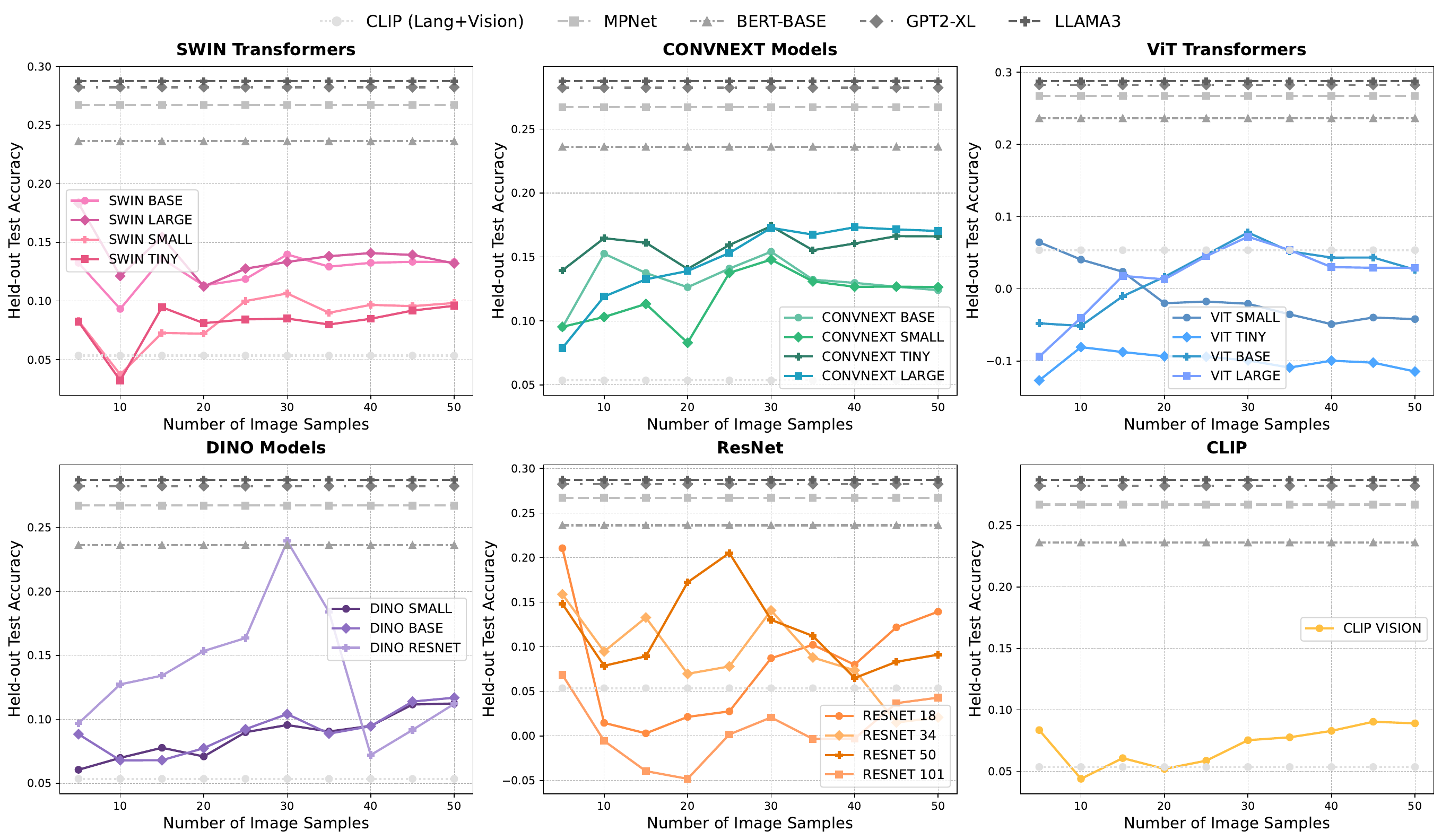}
\caption{Performance comparison between LLM embeddings of the original linguistic stimuli from the Tuckute (2024) dataset and vision model embeddings of their corresponding visual counterparts.}
\label{fig:greta_img_sent}
\end{figure}

\begin{figure}[h!]
\centering
\includegraphics[width=\linewidth]{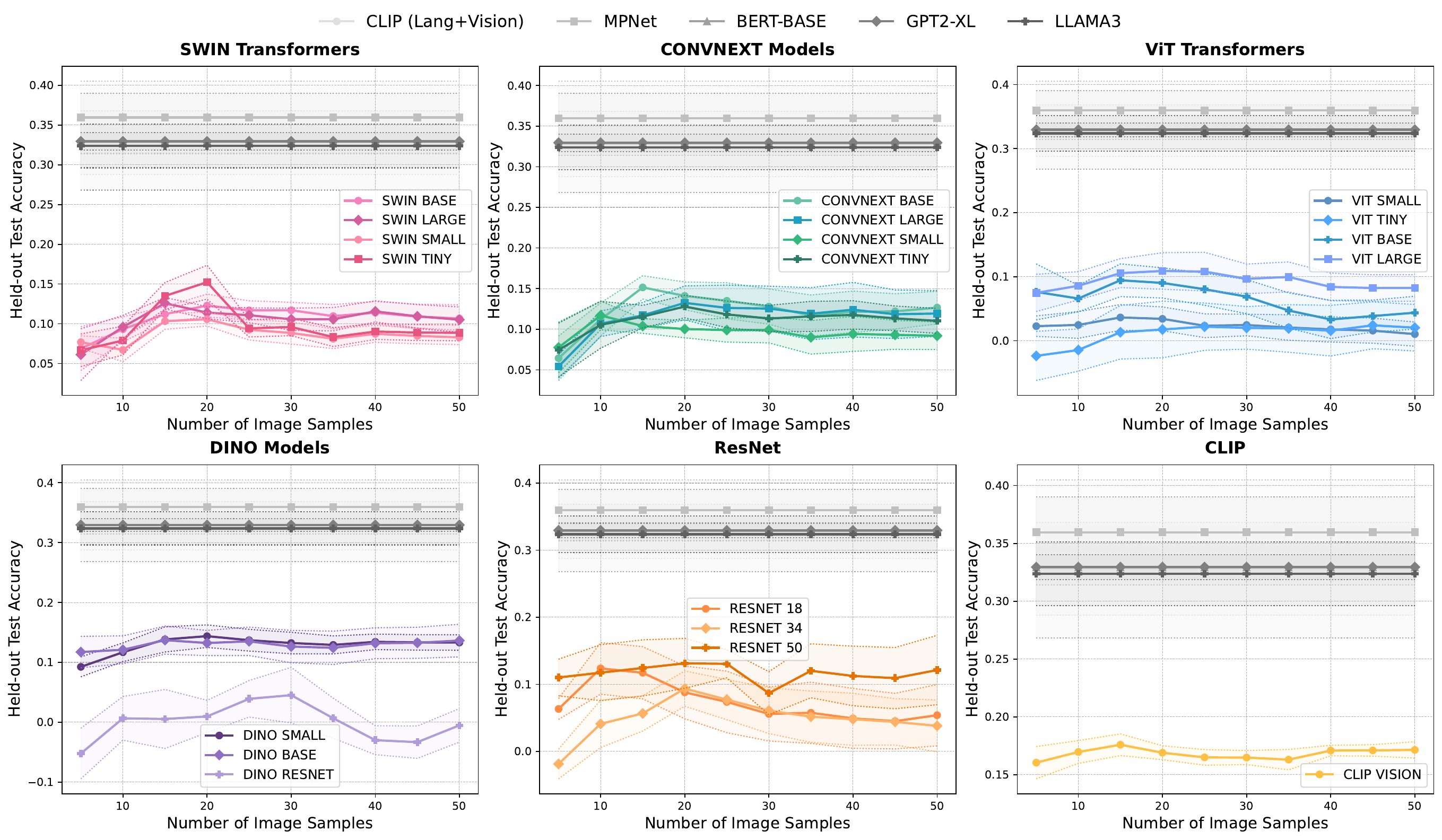}
\caption{Evaluating LLM embeddings of Tuckute (2024) sentences versus vision-model embeddings of their visual counterparts, using only sentences with generated images of sufficient quality (mean CLIP score $\ge$ 0.25)}
\label{fig:greta_img_sent_good_clip}
\end{figure}

% Among the various vision models evaluated, the best-performing ones usually belong to the Swin Transformer family, followed closely by the ConvNeXt models. Both architectures draw inspiration from the strengths of convolutional neural networks (CNNs) and vision transformers (ViTs), blending local feature extraction with elements of global context modeling. Swin Transformers stand out due to their use of shifted windows, which allow for fine-grained, hierarchical feature learning with localized attention, while also incorporating global, non-local interactions through cross-window connections. This combination enables Swin models to capture both detailed spatial structure and broader semantic context more effectively than standard ViTs or convolutional networks. ConvNeXt models, on the other hand, are CNN-based architectures that integrate several transformer-inspired design principles, such as layer normalization and large kernel convolutions, to enhance their global receptive field. Although they do not employ explicit attention mechanisms, their design choices—like enlarged pooling kernels and improved normalization—approximate global feature integration in a more parameter-efficient way. Despite these advances, their ability to model long-range dependencies appears somewhat weaker compared to Swin Transformers, which may explain the performance gap observed in our evaluations.

\subsection{Datasets}
\label{subsec:datasets}

The following datasets were used in our study: 
\begin{enumerate}
    \item \cite{pereira2018toward}, who recorded brain responses from 16 native English speakers as they silently read short, syntactically and semantically diverse passages. The study comprised three experiments: (1) isolated words and simple phrases to probe basic lexical–semantic representations, (2) 384 full sentences grouped into 96 short narratives on varied concrete topics such as professions, clothing, birds, musical instruments, natural disasters, and crimes, and (3) an additional 243 full sentences organized into 72 narratives of similar thematic breadth. For our analyses, we focus on the five participants who completed both Experiments 2 and 3, using only these two experiments because they provide full-length sentence stimuli: 627 unique sentences in total, suitable for modeling rich compositional semantics.
  
    \item \cite{tuckute2024driving}, who recorded brain responses from five native English speakers as they read 1,000 six-word sentences in an event-related fMRI design. The sentences were sampled from text corpora to maximize both semantic coverage and stylistic diversity. Each sentence was presented individually (2 s on screen, 4 s inter-stimulus interval), with runs of 50 sentences and short fixation blocks interleaved. Participants were instructed to read attentively and think about the meaning of each sentence, and to encourage engagement, they were informed that a brief memory task would follow the scan. Because each stimulus was presented to participants only once, we trained the encoding models to predict the average response across all voxels in the functionally-defined five language fROIs. This dataset provides brain responses to a semantically and stylistically diverse set of decontextualized sentences, well suited for modeling sentence-level representations. 
    
    Note that the sentences in this dataset are relatively abstract and simpler compared to those in the Pereira (2018) dataset. As a result, many sentences lack clearly identifiable content words (see Section \ref{subsec:training_dataset}). For this reason, we did not perform the content-word experiments on this dataset.
    
    \item \cite{li2025large} (Caption Scene Dataset) comprises fMRI data from eight participants performing a semantic matching task: each participant first read a Chinese caption and then viewed a corresponding MS-COCO image \cite{lin2014microsoft} to decide whether the text and image matched semantically. For our analyses, we focused on the four subjects with the highest signal-to-noise ratios. Because the image was shown only after the caption, the neural responses during caption reading are uncontaminated by visual input. We therefore analyze only these “pure language” responses from the caption-reading phase. The full dataset contains 9,375 unique captions and 9,494 images (18,893 total stimuli). From these, we use the 983 caption stimuli that were presented to all four selected subjects. The accuracies reported on this dataset are noise-normalized, with noise ceiling computed following the NSD procedure \cite{allen2022massive}. In our case, given only two repetitions for each stimulus, the noise ceiling is equivalent to split-half reliability.
    Note that since the captions in the Captions Scene Dataset are written in Chinese, we do not apply our linguistic modulation analyses, such as isolating content words or generating paraphrases to this dataset.  
\end{enumerate}

\subsection{Training Details}
\label{subsec:training_details}

For the Pereira (2018) dataset, we trained the model using backpropagation with a combined mean squared error and correlation loss between the true and predicted voxel responses. This is because a few voxel measurements contained \texttt{NaN} values, we retained all columns and avoided discarding affected columns.  Brain responses for the remaining datasets were estimated using the closed-form ridge solution, with $\lambda$ chosen by $k$-fold cross-validation ($k=1$ for the Tuckute (2024) dataset and $k=10$ for the CSD (2025) dataset).  Test accuracy is quantified as the Pearson correlation coefficient computed between the predicted and observed voxel responses across the held-out test set.

When training via backpropagation, we cross-validated across a grid of $\lambda$ values ($0.0,0.1,0.01,0.001,0.0001$). Further, for each dataset we generated multiple train–validation–test splits (3 for Pereira (2018), and 5 for the rest) and reported the averaged results across them. For the Pereira (2018) dataset, we designed these splits to ensure a fair evaluation, since having sentences from the same paragraph appear in both training and test sets could allow the model to exploit shared context \cite{kauf2024lexical, feghhi2024large}. To assess whether the observed trends hold independently of contextual overlap and temporal autocorrelation, we created two random splits and one targeted variants: a split in which the test set contained only the first sentence from each of 63 paragraphs.

\subsection{Language  Models used for Evaluation}
\label{subsec:lang_details}
\begin{enumerate}
    \item \textbf{CLIP Language Encoder} \cite{radford2021learning}:  
    The text branch of CLIP, trained jointly with its vision counterpart using a contrastive image–text objective. It produces rich sentence-level embeddings aligned to visual concepts, enabling zero-shot cross-modal tasks.

    \item \textbf{MPNET} \cite{song2020mpnet}:  
    A Transformer-based language model that combines masked language modeling with permutation-based training, allowing it to capture bidirectional context and sequential dependency simultaneously.

    \item \textbf{BERT Base} \cite{devlin2019bert}:  
    A bidirectional Transformer pretrained with masked language modeling, providing contextual word and sentence embeddings.

    \item \textbf{GPT-2 XL} \cite{radford2019language}:  
    An autoregressive Transformer with 1.5 B parameters, trained to predict the next token in large-scale web text.

    \item \textbf{Gemma-3} \cite{team2025gemma}:  
    Google’s latest open-weight decoder-only Transformer with grouped-query attention and alternating local/global layers for efficient long-context reasoning (up to 128K tokens). Larger variants add a SigLIP vision encoder for multimodal inputs and support quantization for lightweight deployment.

    \item \textbf{LLaMA-3} \cite{dubey2024llama}:  
    Meta’s third-generation decoder-only Transformer featuring grouped-query attention and efficient scaling.
\end{enumerate}

\subsection{Vision Models used for Evaluation}

\begin{enumerate}
    \item \textbf{ResNet family} \cite{he2016deep} (ResNet-50, ResNet-34, ResNet-18, ResNet-101):  
    Deep residual networks that use skip connections to mitigate vanishing gradients, enabling very deep convolutional architectures. 

    \item \textbf{ConvNeXt family} (ConvNeXt Small, Tiny, Base, Large) \cite{liu2022convnet}:  
    A modernized CNN design that incorporates architectural ideas from Transformers, such as large kernels and inverted bottlenecks. 

    \item \textbf{Swin Transformers} \cite{liu2021swin} (Swin Small, Tiny, Base, Large):  
    Hierarchical vision Transformers that process images using shifted windows, providing linear computational complexity with respect to image size.

    \item \textbf{Vision Transformers (ViTs)} (ViT Small, Tiny, Base, Large) \cite{dosovitskiy2020image}:  
    Pure Transformer architectures that treat images as sequences of non-overlapping patches, capturing global context through self-attention.

    \item \textbf{DINO models} (DINO Small, Base, ResNet-50) \cite{caron2021emerging}:  
    Self-supervised representations learned via knowledge distillation, producing strong, transferable visual features without labeled data and supporting both ViT and ResNet backbones.

    \item \textbf{CLIP Vision Encoder} \cite{radford2021learning}:  
    The visual branch of CLIP, trained with a contrastive objective to align images and text in a shared embedding space, enabling zero-shot recognition and robust cross-modal retrieval.
\end{enumerate}

\label{subsec:vision_details}
\subsection{Evaluating Vision- and Language-Model Predictions of Language Cortex Activity - Tuckute (2024)}

Applying the same vision-versus-language modeling framework to the Tuckute (2024) dataset as in Section \ref{subsec:img_sent}, we do not observe the clear performance gains with increasing image counts seen for Pereira (2018). Many vision models fail to learn meaningful mappings, as indicated by their negative test-time correlation scores (Figure \ref{fig:greta_img_sent}). This likely reflects the more abstract nature of the Tuckute (2024) sentences, which makes generating semantically aligned visuals more challenging (Figure \ref{fig:Pereira_Dataset_Samples}). Nevertheless, the performance of vision-based models remains reasonably competitive, indicating that even abstract sentences can evoke visual representations that capture meaningful information, albeit with more noise. Importantly, this observation still aligns with our core hypothesis: even when vision-based representations are noisier, they retain enough semantic signal to contribute meaningfully to neural prediction.

To account for this variability, we filtered the Tuckute (2024) dataset to retain only those images with a CLIP-based quality score above 0.25. Restricting our analysis to these higher-quality samples allowed us to recover the expected performance trend: as the number of relevant images increases, model accuracy improves. However, this trend is less not observed in ViT-based models (Figure \ref{fig:greta_img_sent_good_clip}). One possible explanation is that standard ViT architectures rely on fixed-size, non-overlapping image patches and lack mechanisms for hierarchical feature extraction or localized inductive biases. As a result, they may be less sensitive to fine-grained variations in image quality or spatial detail, especially in scenarios where the semantic content is subtle or distributed across small regions of the image.

fMRI datasets tailored for multimodal modeling, especially those that align closely with the hypotheses we aim to test are exceptionally rare. In particular, not all datasets contain sentences that are as visually grounded and easily translatable into images as those in the Pereira (2018) dataset. For example, many sentences in the Tuckute (2024) dataset are more abstract in nature, making it substantially more difficult to generate meaningful visual representations. This challenge is evident in the lower quality and occasionally off-topic images produced by the diffusion model, as shown in Appendix Figure \ref{fig:Pereira_Dataset_Samples}.

\subsection{Evaluating Vision- and Language-Model Predictions of Language Cortex Activity (Caption Scene Dataset 2025)}

Lastly, we compare language-model embeddings of the full original sentence with vision-model embeddings of the corresponding visual scenes using the Caption Scene Dataset (2025), as described in Section \ref{subsec:img_sent}. Because participants first read captions describing MS-COCO images, we begin by contrasting the language-model embeddings of these captions with the vision-model embeddings of the original COCO images.

Although the language models better capture brain activity in the language cortex, the vision models achieve performance that is only slightly lower. This reinforces our earlier finding that language cortex prediction remains highly sensitive to the exact linguistic form, giving language models an advantage, while the underlying semantic content can be represented nearly as well through visual modality.

We then extend this analysis by generating multiple synthetic images for each caption with Stable Diffusion and averaging their vision embeddings, following the procedure in Section \ref{subsec:img_sent}. The same pattern emerges as in the Pereira (2018) and Tuckute (2024) datasets (Figures \ref{fig:Pereira_img_sent} and \ref{fig:greta_img_sent_good_clip}): vision-model performance improves as the number of generated samples increases, in some cases approaching that of the language models.

\begin{figure}[h!]
\centering
\includegraphics[width=\linewidth]{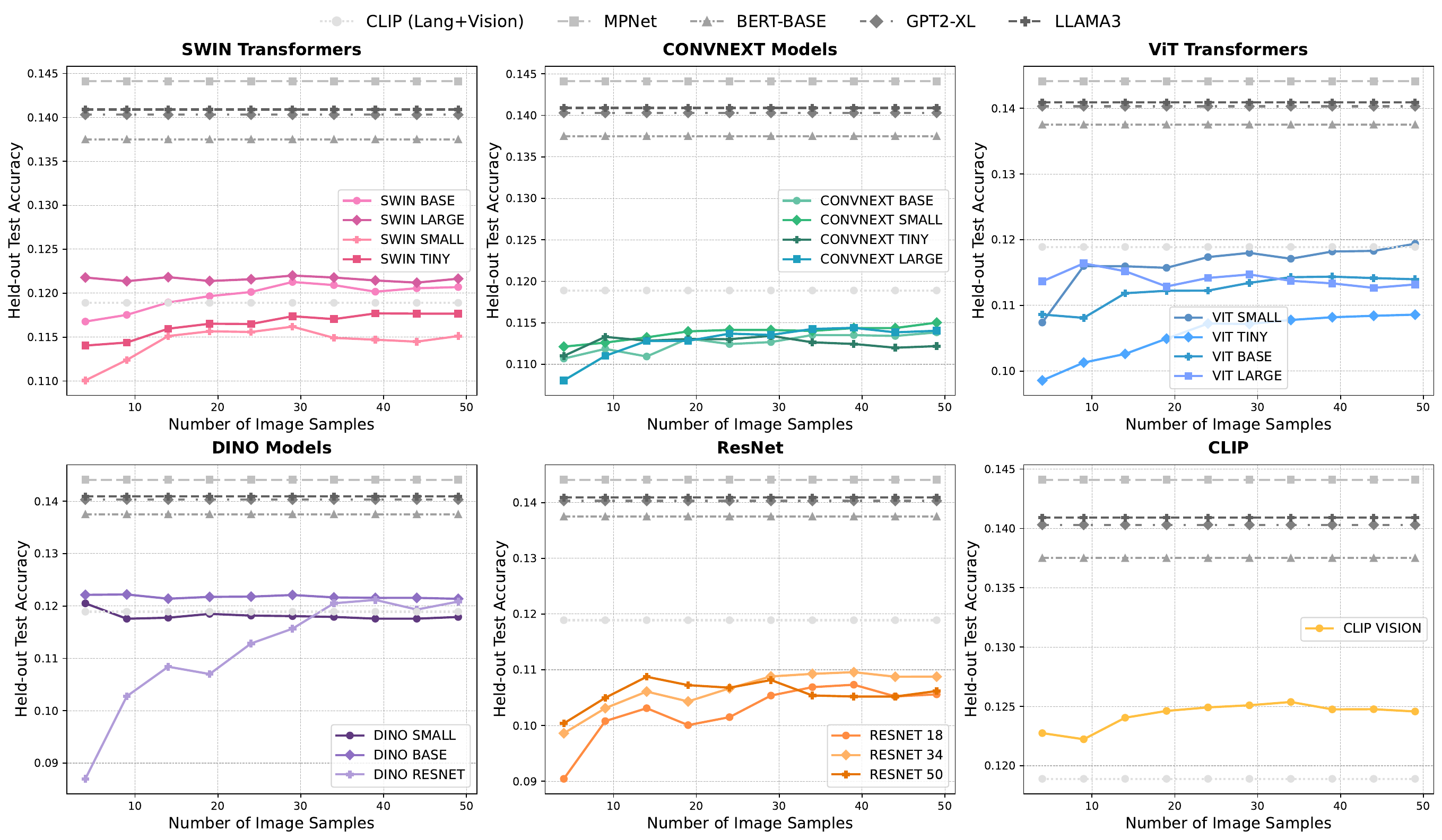}
\caption{Performance comparison between LLM embeddings of the original linguistic stimuli from the CSD (2025) dataset and vision model embeddings of their corresponding visual counterparts.}
\label{fig:csd_img_sent}
\end{figure}

\begin{table}[htbp]
  \centering
  \begin{tcolorbox}
    \centering
    \textbf{Prompt to generate paraphrases}\\[1ex]
    \begin{lstlisting}[basicstyle=\ttfamily\scriptsize,
                       breaklines=true,
                       columns=flexible,
                       frame=none]
Prompt:
f"""You are an expert image captioner. I'll show you some existing captions for an image, and your task is to generate 70 NEW captions that:
    1. Are similar in style and detail level to the existing captions
    2. Capture the same meaning but with different wording
    3. Are direct, concise descriptions (around 10-15 words each)
    4. Are worded differently from each existing caption and from each other

Here are the existing captions:
{insert all captions text for the image here}

Generate 10 new captions formatted exactly as:
1. [First new caption]
2. [Second new caption]
3. [Third new caption]
4. [Fourth new caption]
5. [Fifth new caption]
6. [Sixth new caption]
7. [Seventh new caption]
8. [Eighth new caption]
9. [Ninth new caption]
10. [Tenth new caption] ... """
    \end{lstlisting}
  \end{tcolorbox}
  \caption{Prompt used for generating paraphrase using Gemini-2.5-Flash.}
  \label{tab:para_prompt}
\end{table}

\begin{table}[htbp]
  \centering
  \begin{tcolorbox}
    \centering
    \textbf{Prompt to generate paraphrases with extra context}\\[1ex]
    \begin{lstlisting}[basicstyle=\ttfamily\scriptsize,
                       breaklines=true,
                       columns=flexible,
                       frame=none]
Prompt:
f"""You are an expert image captioner. I'll show you existing captions for an image,
and your task is to generate 70 NEW captions that:

1. Are similar in style but include more detail than the existing captions
2. Capture the same meaning but with different wording
3. Are worded differently from each existing caption and from each other
4. Contain additional commonsense context to the original caption

Example:
For the sentence 'The boy is eating pancakes for breakfast', some paraphrases
with additional context would be:
  1. The boy is eating pancakes with maple syrup in the morning for breakfast
  2. The boy is sitting at the dining table and having pancakes for breakfast

Here are the existing captions:
{insert all captions text for the image here}

Generate 10 new captions formatted exactly as:
1. [First new caption]
2. [Second new caption]
3. [Third new caption]
4. [Fourth new caption]
5. [Fifth new caption]
6. [Sixth new caption]
7. [Seventh new caption]
8. [Eighth new caption]
9. [Ninth new caption]
10. [Tenth new caption] ... """
    \end{lstlisting}
  \end{tcolorbox}
  \caption{Prompt used for generating paraphrases with additional commonsense context using Gemini-2.5-Flash.}
  \label{tab:para_context_prompt}
\end{table}

\begin{figure}[h!]
\centering
\includegraphics[width=0.9\linewidth]{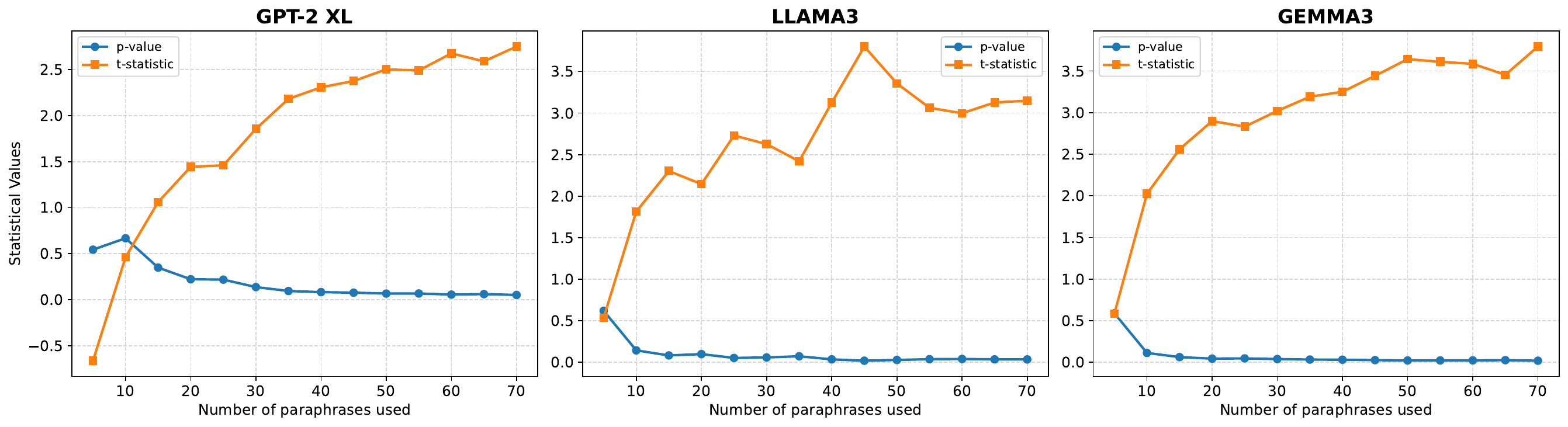}
\caption{Expanded analysis of the experiments described in Section \ref{subsec:para_details} across five data splits. For each condition, we compute test-time accuracies for averaged paraphrases and compare them with the accuracies obtained using the original single sentence. The t-statistics increase with the number of paraphrases, further supporting the claims in Section \ref{subsec:para_details}.}
\label{fig:stat_analysis_paraphrases}
\end{figure}

\begin{figure}[h!]
\centering
\includegraphics[width=0.9\linewidth]{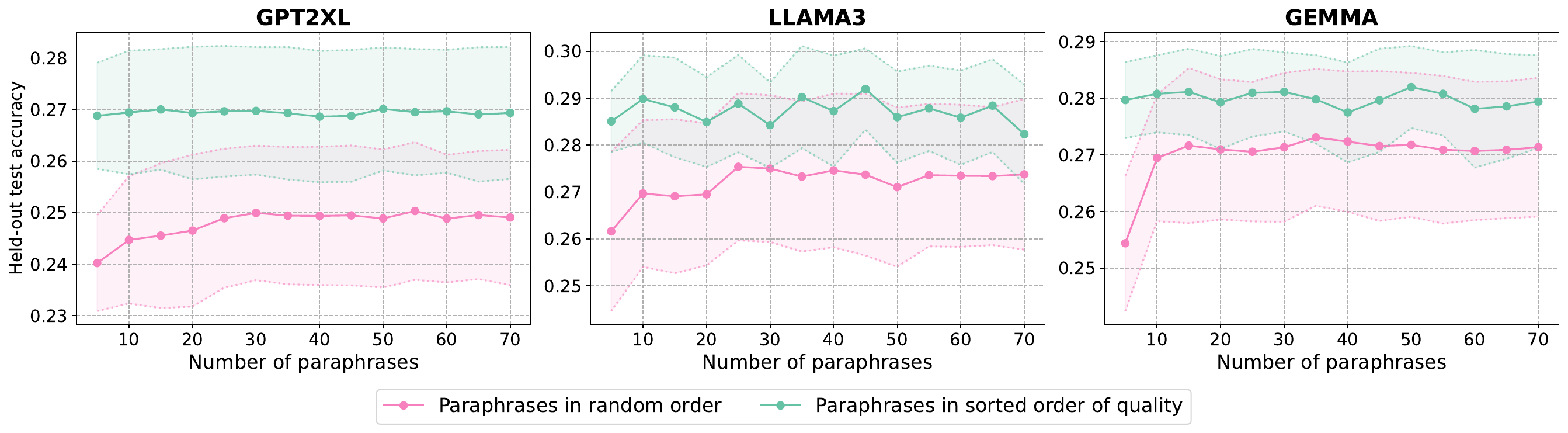}
\caption{Pereira (2018) dataset - Comparison of averaged LLM embeddings of paraphrases in random order with those arranged in sorted order of semantic similarity to the original sentence. These paraphrases do not have added commonsense context.}
\label{fig:paraphrase_no_pereira_rand_sort}
\end{figure}

\begin{figure}[h!]
\centering
\includegraphics[width=0.9\linewidth]{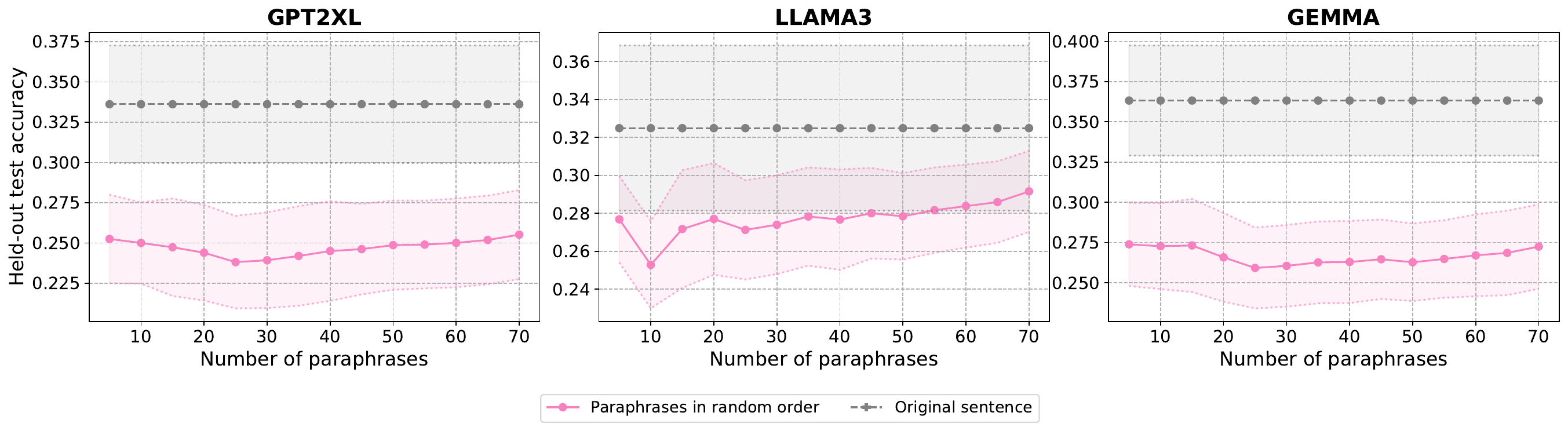}
\caption{Tuckute (2024) dataset - Comparison of LLM embeddings of the original linguistic stimuli presented to subjects with averaged embeddings of generated paraphrases without additional context.}
\label{fig:paraphrase_no_tuckute_org}
\end{figure}

\begin{figure}[h!]
\centering
\includegraphics[width=0.9\linewidth]{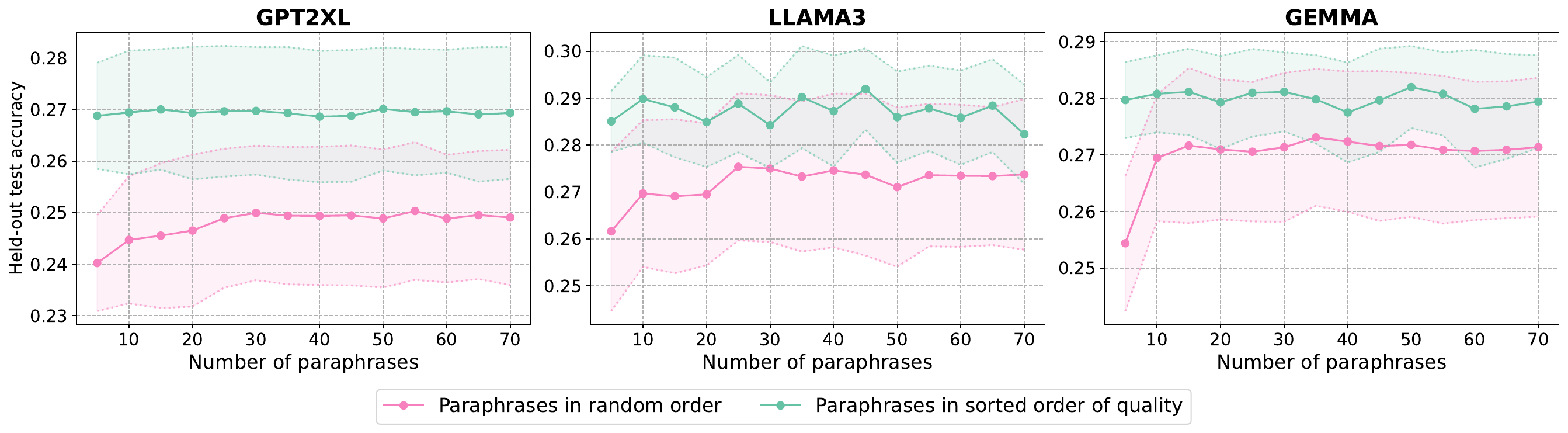}
\caption{Pereira (2018) dataset - Comparison of averaged LLM embeddings of paraphrases in random order with those arranged in sorted order of semantic similarity to the original sentence. These paraphrases do not have added commonsense context.}
\label{fig:paraphrase_no_pereira_rand_sort}
\end{figure}

\begin{figure}[h!]
\centering
\includegraphics[width=0.9\linewidth]{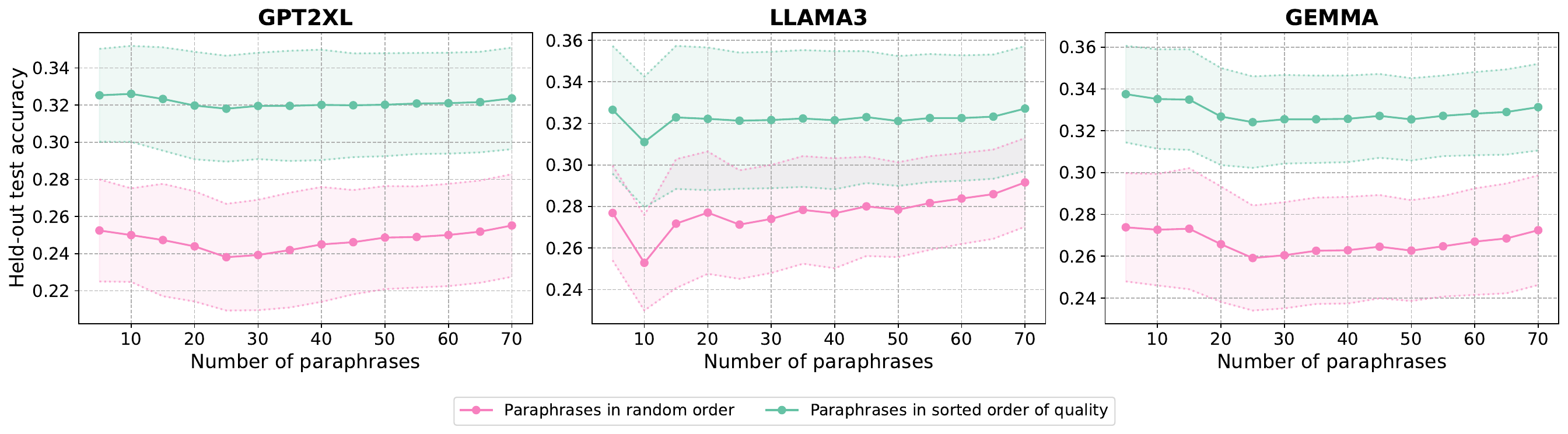}
\caption{Tuckute (2024) dataset - Comparison of averaged LLM embeddings of paraphrases in random order with those arranged in sorted order of semantic similarity to the original sentence. These paraphrases do not have added commonsense context.}
\label{fig:paraphrase_no_tuckute_rand_sort}
\end{figure}

\begin{figure}[h!]
\centering
\includegraphics[width=0.9\linewidth]{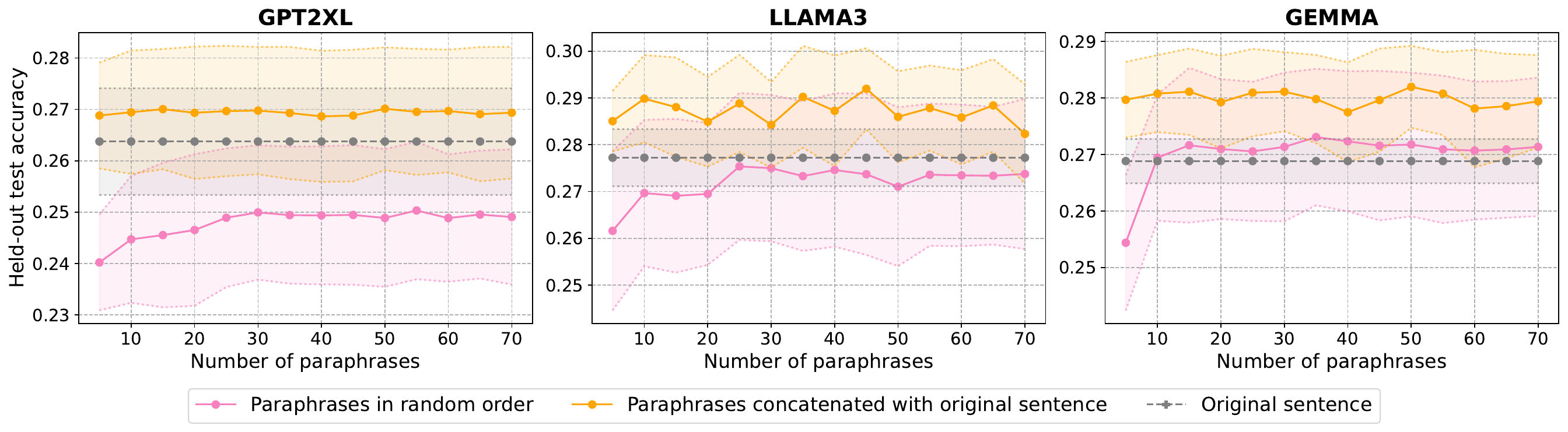}
\caption{Pereira (2018) dataset - Comparison of averaged LLM embeddings of paraphrases alone with those that are concatenated with the original sentence. These paraphrases do not have added commonsense context.}
\label{fig:paraphrase_no_pereira_rand_conc}
\end{figure}

\begin{figure}[h!]
\centering
\includegraphics[width=0.9\linewidth]{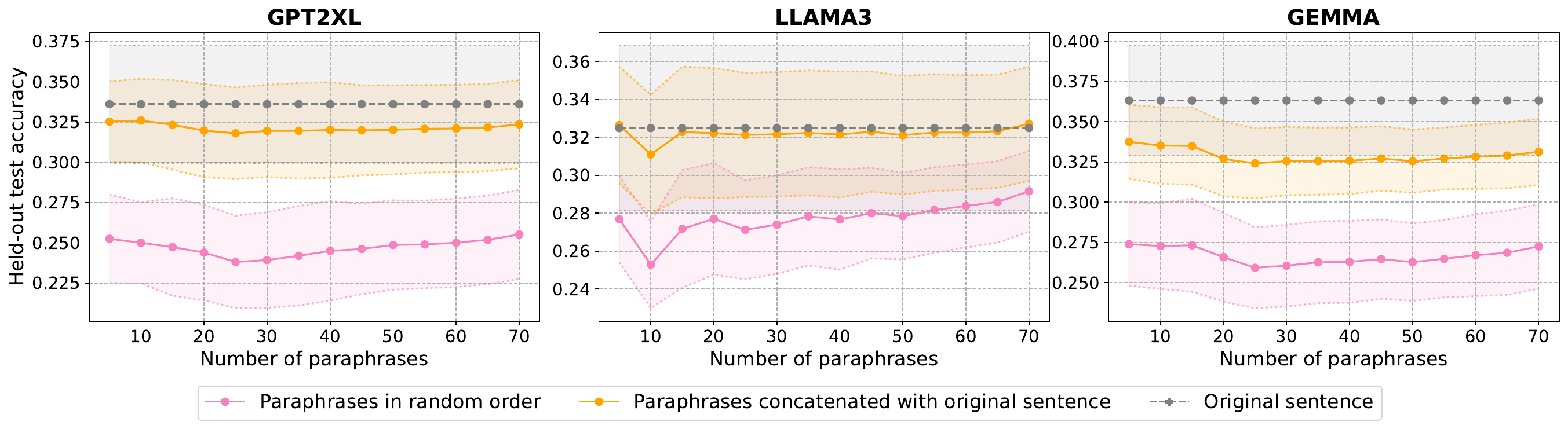}
\caption{Tuckute (2024) dataset - Comparison of averaged LLM embeddings of paraphrases alone with those that are concatenated with the original sentence. These paraphrases do not have added commonsense context.}
\label{fig:paraphrase_no_tuckute_rand_conc}
\end{figure}

\begin{figure}[h!]
\centering
\includegraphics[width=0.9\linewidth]{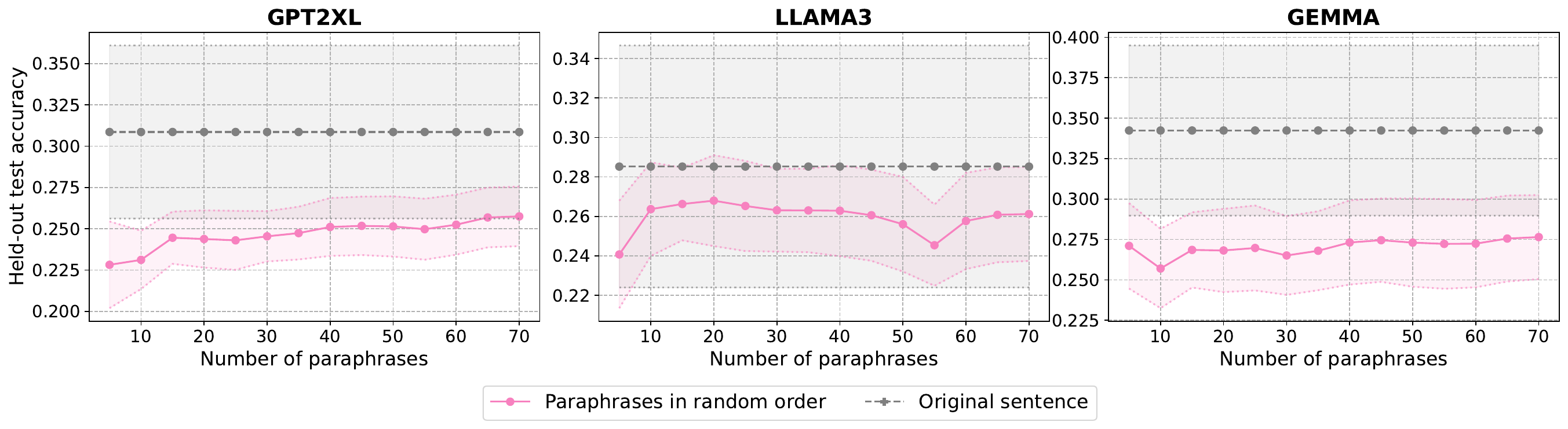}
\caption{Tuckute (2024) dataset - Comparison of LLM embeddings of the original linguistic stimuli presented to subjects with averaged embeddings of generated paraphrases with additional context.}
\label{fig:paraphrase_extra_tuckute_org}
\end{figure}

\begin{figure}[h!]
\centering
\includegraphics[width=0.9\linewidth]{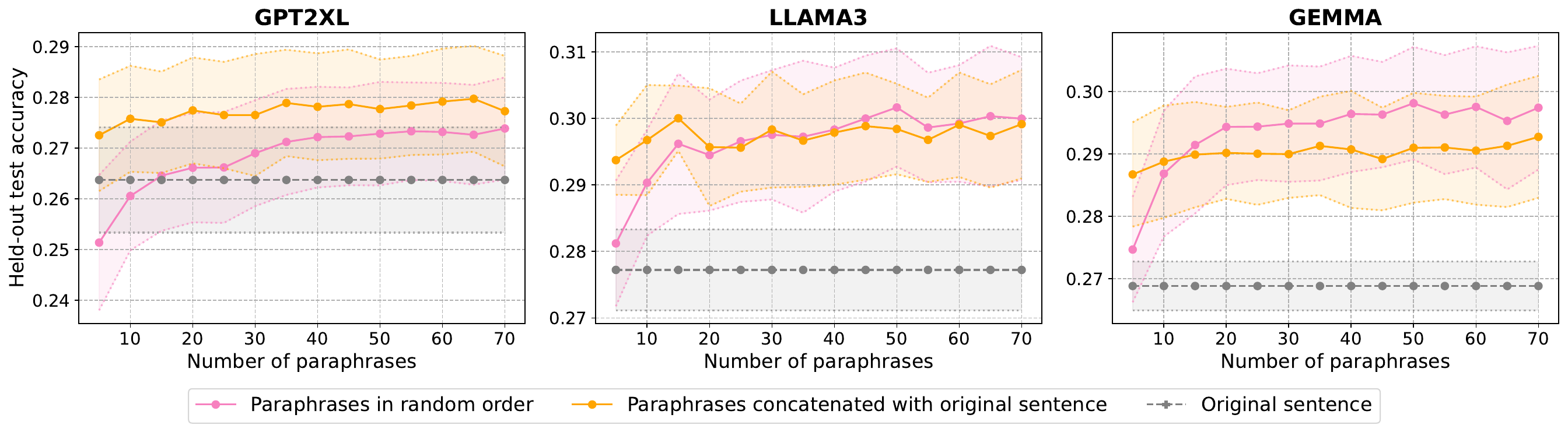}
\caption{Pereira (2018) dataset - Comparison of averaged LLM embeddings of paraphrases alone with those that are concatenated with the original sentence. These paraphrases have added commonsense context.}
\label{fig:paraphrase_extra_pereira_rand_conc}
\end{figure}

\begin{figure}[h!]
\centering
\includegraphics[width=0.9\linewidth]{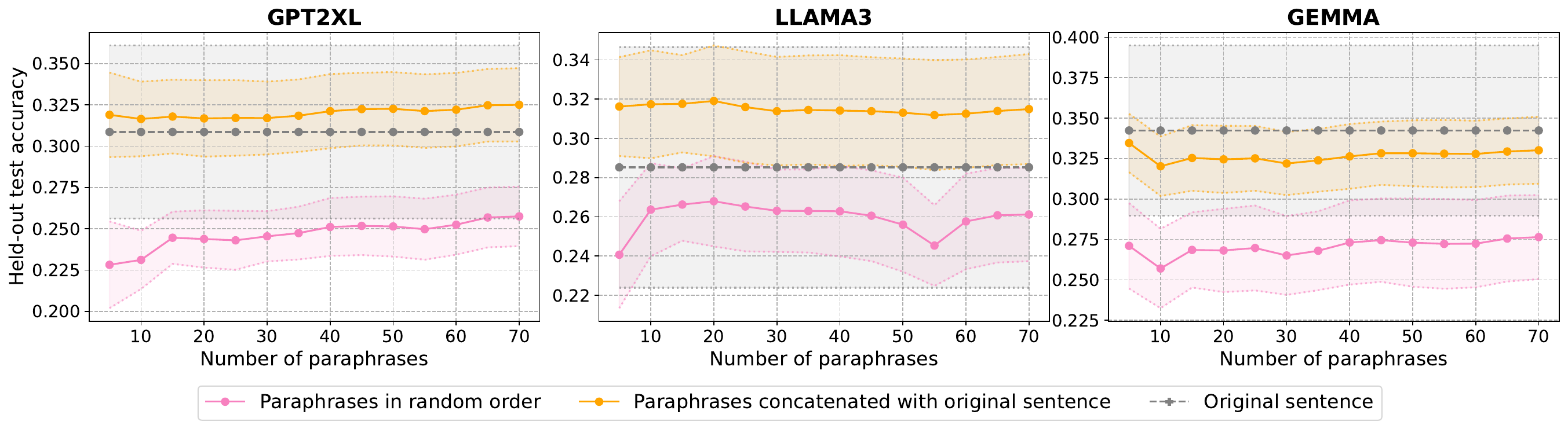}
\caption{Tuckute (2024) dataset - Comparison of averaged LLM embeddings of paraphrases alone with those that are concatenated with the original sentence. These paraphrases have added commonsense context.}
\label{fig:paraphrase_extra_tuckute_rand_conc}
\end{figure}

\subsection{THE USAGE OF LARGE LANGUAGE MODELS (LLMS)}

LLMs are primarily used to
identify typos and make the language more aligned with conventions of academic writing.

\end{document}